\documentclass[a4paper,fleqn]{cas-dc}

\usepackage[numbers]{natbib}

\usepackage[font=footnotesize]{caption}
\DeclareCaptionLabelFormat{table}{Table~\thetable}
\usepackage{subcaption}

\usepackage{booktabs}
\usepackage{adjustbox}
\usepackage{multirow}
\usepackage{tikz-dependency}

\usepackage{ltablex}

\usepackage{placeins}
\usepackage{float}
\usepackage{caption}

\usepackage[dvipsnames]{xcolor}

\newcommand{\imran}[1]{\textcolor{black}{#1}}

\definecolor{lightpurple}{HTML}{E1D5E7}
\definecolor{purple}{HTML}{9673A6}
\definecolor{lightgreen}{HTML}{D5E8D4}
\definecolor{green}{HTML}{82B366}
\definecolor{blue}{HTML}{6C8EBF}
\definecolor{lightblue}{HTML}{DAE8FC}

\usepackage[most]{tcolorbox}
\tcbset{
  coltext=black,     
  colframe=black,    
  boxrule=0.2mm,     
  arc=0.5mm,           
  left=0.1mm, right=0.1mm, top=0mm, bottom=0mm, 
  on line,
}
\newcommand{\purplebox}[1]{\tcbox[colback=lightpurple, colframe=purple]{\footnotesize#1}}
\newcommand{\greenbox}[1]{\tcbox[colback=lightgreen, colframe=green]{\footnotesize#1}}
\newcommand{\bluebox}[1]{\tcbox[colback=lightblue, colframe=blue]{\footnotesize#1}}

\def\tsc#1{\csdef{#1}{\textsc{\lowercase{#1}}\xspace}}
\tsc{WGM}
\tsc{QE}
\tsc{EP}
\tsc{PMS}
\tsc{BEC}
\tsc{DE}

\begin{document}
\let\WriteBookmarks\relax
\def\floatpagepagefraction{1}
\def\textpagefraction{.001}

\shorttitle{Structured Sentiment Analysis}

\shortauthors{Imran et~al.}

\title [mode = title]{Structured Sentiment Analysis Using Sequence Labeling as Dependency Graph Parsing}                      



\author[1]{Muhammad Imran}[
                        orcid=0000-0001-7511-2910
                        ]

\cormark[1]


\ead{m.imran@udc.es}


\credit{Conceptualization, Data curation, Formal analysis, Investigation, Methodology, Software, Validation, Visualization, Writing - original draft, Writing - review \& editing}

\affiliation[1]{organization={Universidade da Coruña, CITIC, Departamento de Ciencias de la Computación y Tecnologías de la Información},
    addressline={Campus de Elviña s/n}, 
    postcode={15071}, 
    city={A Coruña},
    country={Spain}}
    
\author[2]{Ana Ezquerro}[]
\credit{Conceptualization, Methodology, Software, Validation, Visualization, Writing - original draft, Writing - review \& editing}
\affiliation[2]{organization={Graz University of Technology, Institute in Machine Learning and Neural Computation},
    addressline={Rechbauerstraße 12}, 
    postcode={8010}, 
    city={Graz},
    country={Austria}}

\author[1]{Carlos Gómez-Rodríguez}[%
   ]

\credit{Conceptualization, Formal analysis, Funding acquisition, Investigation, Methodology, Project administration, Supervision, Validation, Writing - original draft, Writing - review \& editing}

\author%
[3]
{Anders Søgaard}
\credit{Conceptualization, Investigation, Methodology, Supervision}
\affiliation[3]{organization={University of Copenhagen, Department of Computer Science},
    addressline={Lyngbyvej 2}, 
    postcode={DK-2100}, 
    city={Copenhagen},
    country={Denmark}}

\author%
[1]
{David Vilares}

\credit{Conceptualization, Formal analysis, Investigation, Methodology, Validation, Writing - original draft, Writing - review \& editing}
\cortext[cor1]{Corresponding author}



\begin{abstract}
This study addresses the problem of structured sentiment analysis, whose goal is to obtain a fine-grained sentiment graph where the nodes represent spans of sentiment holders, targets, and expressions, while the arcs define the relationships among them. Our proposed approach casts the task as dependency graph parsing, but departs from traditional parsing methods by solving it through sequence labeling. To do so, we leverage recent advances in linearized graph encodings that allow each word in the input to be assigned a label, effectively capturing the structure of the dependency graph. We conducted experiments on seven datasets spanning five languages (English, Spanish, Norwegian, Basque, and Catalan), showing performance competitive with leading, more complex single-model approaches.
\end{abstract}




\maketitle

\section{Introduction} \label{introduction}
Traditional approaches to Sentiment Analysis (SA) have largely
focused on relatively simple tasks like sentence- or document-level polarity classification, predicting coarse sentiment categories based on the overall tone of a text (e.g., positive, negative, or neutral) \cite{pang-etal-2008-opinion, cui-etal-2023-survey}. However, these methods often fail to capture the nuanced structure of sentiment in complex linguistic expressions, such as those involving negation, ambiguity, or multiple sentiment targets.
To overcome these limitations, more fine-grained approaches have been investigated. For instance, a substantial body of research has explored aspect-based sentiment analysis (ABSA) \cite{pontiki-etal-2014-semeval,schouten2015survey}, where sentiment polarities are associated with specific entities or aspects of a text rather than documents or sentences as a whole. In turn, Structured Sentiment Analysis (SSA) takes an even more fine-grained approach which deals with SA as a structured information extraction problem in the shape of a  sentiment graph, aiming to uncover not only sentiment polarity but also the underlying structure of who expresses what sentiment, about what, and how these components interact.

\begin{figure}
	\centering
    \resizebox{\columnwidth}{!}{%
		\includegraphics[scale=.75]{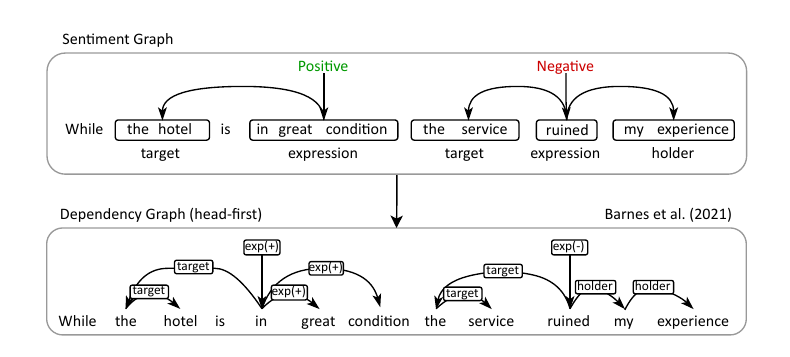}%
    }
	\caption{Head-first transformation of a sentiment graph to a dependency graph \cite{barnes-etal-2021-structured}. The sentiment graph has two opinion tuples: $O_1=(\emptyset$, \textit{the hotel}, \textit{in great condition}, \texttt{+}$)$ and $O_2=($\textit{my experience}, \textit{the service}, \textit{ruined}, \textit{-}$)$.}
	\label{fig:ssa-graph}
\end{figure}

Following this line, \citet{barnes-etal-2021-structured}
formalize SSA as the task of identifying all opinion tuples 
$\{O_1, \dots, O_{m} \}$ in a given text $W=(w_1,...,w_n)\in\mathcal{V}^n$.\footnote{From now on, we use the symbol $\mathcal{V}$ to denote the vocabulary of tokens, $\mathcal{P}$ for the set of sentiment polarities, and $n$ and $m$ for the number of tokens and opinion tuples, respectively, in an input text.} Each opinion tuple is denoted as $O_i=(h,t,e,p)$, where \(h\subset W\) is the \textit{opinion holder} who expresses a \textit{polarity} \(p\in\mathcal{P}\) towards a \textit{target} \(t\subset W\) through some sentiment \textit{expression} \(e\subset W\), implicitly establishing pairwise relationships among elements of the same tuple (Figure \ref{fig:ssa-graph}). These elements are pivotal to completely resolve the sentiment analysis problem \cite{liu-2012-sentiment} as they are highly interconnected.

Despite the progress in SA, traditional approaches---including lexicon-based techniques and early supervised machine learning models---often struggle to handle the complexity of structured sentiment information \cite{pontiki-etal-2016-semeval}. Lexicon-based methods~\cite{liu-2012-sentiment,taboada2011lexicon}, while interpretable and easy to implement, are brittle when confronted with syntactic variations and phenomena such as negation. Meanwhile, supervised learning and neural network-based approaches have achieved strong results in coarse-grained tasks, but generally treat sentiment as a classification problem, overlooking the complex relationships between holders, targets, expressions, and polarities; and lacking explainability \cite{ribeiro2016should, zhang2018deep}. In contrast, SSA explicitly models these interdependencies, offering a more comprehensive and interpretable view of sentiment in context.

The accurate and efficient extraction of sentiment structures remains an open challenge in SSA. Recent advances in dependency parsing and semantic graph-based representations provide promising directions in this respect~\cite{barnes-etal-2021-structured, peng2021sparse}. These approaches leverage semantic parsing algorithms to better capture relational dependencies among sentiment elements, leading to more interpretable and accurate sentiment classification. In this work, we approach SA as a dependency graph parsing problem, where the output is a directed acyclic graph. Sentiment expressions act as source nodes, while the remaining elements are connected through arcs that represent their relationships. This approach encourages us to exploit the current improvements in semantic dependency parsing \cite{dozat-manning-2018-simpler, oepen-etal-2020-mrp, kurtz-etal-2020-end, samuel-straka-2020-ufal} to build an efficient semantic graph parser capable of predicting the structured sentiment graph as shown in Figure~\ref{fig:ssa-graph}. 
However, existing SSA approaches based on dedicated graph parsers often rely on complex architectures and specialized decoding procedures \cite{morio-etal-2022-hitachi}. This motivates exploring simpler yet expressive alternatives that can leverage the strengths of pretrained language models while preserving the structured nature of sentiment graphs.

In this study, we aim to contribute to the field of structured sentiment analysis by exploiting advances in dependency graph
parsing models. Our approach is grounded in the observation that SSA graphs exhibit specific structural properties—such as sparsity, bounded dependency distances, and a limited set of relation types—that make them well suited to linearized dependency encodings.  For this purpose, we use the recently-developed encodings that, in turn, cast dependency graph parsing as a sequence labeling task~\cite{ezquerro-etal-2024-dependency}. This allows us to solve the SSA problem in an efficient and straightforward way.
We evaluate these models on seven datasets across five different languages, focusing on their ability to capture the inter-dependencies between opinion tuples and their performance in monolingual settings. In our experiments, we demonstrate that 
sequence-labeling dependency graph parsing models,
when combined with pretrained language models, can obtain strong results on various datasets. 
Ultimately, this work establishes that sequence labeling, when governed by the structural constraints of SSA, provides a principled and theoretically well-founded decoding strategy for sentiment graph prediction, resulting in a particularly simple and efficient approach to the task.

\section{Related Work}\label{related-work}

Sentiment analysis has long been a popular task in natural language processing (NLP), with approaches evolving from lexicon-based and machine learning methods \cite{taboada2011lexicon, maas-etal-2011-learning} to modern deep learning architectures \cite{tian-etal-2020-skep, barbieri-etal-2022-xlm}. In its early stages, sentiment analysis was typically framed as a classification task, where systems were trained to extract sentiment information from raw text at the sentence or document level. The sentiment output was commonly expressed as a polarity (positive, neutral, negative),  often conveyed through a numerical rating scale (e.g., 1 to 5 stars). Alternatively, sentiment could also be represented as a categorical emotion~\cite{strapparava-mihalcea-2007-semeval}, such as anger, joy, or fear.

Initial approaches relied on manual features such as lexicons \cite{taboada2011lexicon, bakliwal-etal-2013-sentiment, mohammad-etal-2013-nrc} to identify the polarity of an input text. As machine learning became more popular in NLP, the value of data-driven features became evident, 
leading to new systems based on probabilistic optimization \cite{yang-cardie-2014-context} or supervised learning \cite{mullen-collier-2004-sentiment, pang-etal-2002-thumbs}.
In this context, the incorporation of syntactic knowledge showed a substantial influence to understand complex linguistic structures 
\cite{poria-etal-2014-sentic,Vilares2017,gomez2024dancing,imran2026syntax}. 
Neural networks have reshaped sentiment analysis by enabling end-to-end learning with dense, distributed representations. This is thanks to models ranging from recurrent to modern Transformer-based language models, such as masked pretrained language models (MLMs), which remove the need for hand-crafted features.

Despite its practical usefulness and the improvements in accuracy achieved with modern architectures, this sentence- or document-level view of SA
still lacks in expressiveness when extracting sentiment information from natural language. It is often the case that a text of interest contains differing opinions on different aspects of an entity, or contrasting opinions from different sources, which cannot be captured by framing the problem as classification of the text as a whole.
Fine-grained formulations have evolved from ABSA to the more recent SSA, addressing this limitation by explicitly identifying the sentiment-bearing elements and the relationships between them, thereby enabling a more detailed representation of opinions.

\subsection{Aspect-based sentiment analysis}

In the domain of ABSA, recent research has explored different architectural innovations to improve fine-grained sentiment extraction. \citet{wu2024lsoit} introduced a lexicon and syntax enhanced opinion induction tree model that dynamically builds a tree with the target aspect at its root. First, the model integrates sememe knowledge, phrase structures, and dependency relations. An attention-based reinforcement learning agent then prunes the tree to maximize sentiment rewards. Finally, a graph convolutional network generates aspect representations for polarity classification. Building on the importance of local context, 
\citet{zhu2025aspect} encode text as spiking-neuron sequences and process sentences bidirectionally to capture surrounding opinion cues. The model separates context and aspect features using a two-channel architecture. It then applies dual attention to dynamically link both representations, assigning higher weights to sentiment-critical words around the target aspect for polarity classification. 
In a complementary line of work, \citet{dubey2025adaptive} proposed an adaptive contextual memory graph transformer. The model integrates (i) a domain-adaptive knowledge graph to capture sentiment shifts across domains, (ii) a hierarchical graph encoder to model multi-relational word-aspect dependencies, and (iii) contrastive learning with semantic role labeling to detect implicit aspects. 
Along similar lines, \citet{gu2025information} proposed an information-assisted multitask learning network that anchors on aspect positions via boundary-aware attention. The model maps sentiment relationships and causes, fuses semantic, syntactic, and lexicon-based representations, and jointly learns aspect extraction and sentiment classification. It further incorporates commonsense knowledge from ConceptNet \cite{liu2004conceptnet} to preserve global context while retaining local granularity.
 
While ABSA research has traditionally been driven by MLMs \cite{li-etal-2019-exploiting} combined with additional neural modules \cite{zhang-etal-2019-aspect} or rich syntactic features \cite{huang-etal-2020-syntax} to enhance contextualization, recent work has diversified into novel paradigms such as lexicon and syntax-enhanced opinion tree induction with reinforcement learning \cite{wu2024lsoit}, spiking-neuron sequence encoding with dual attention \cite{zhu2025aspect}, an adaptive contextual memory graph transformer with contrastive learning \cite{dubey2025adaptive}, and an information-assisted multitask learning network with boundary-aware attention and commonsense knowledge \cite{gu2025information}. 

SSA instead gained popularity with its formalization
in the SemEval 2022 Shared Task 10 \cite{barnes-etal-2022-semeval}. Since then, it has often been reframed as a dependency graph parsing task \cite{barnes-etal-2021-structured, peng2021sparse, samuel-etal-2022-direct, fernandez2026structured}, enabling the application of dependency and semantic parsing techniques, which we detail next.

\subsection{Structured Sentiment Analysis}

Early approaches to SSA focused on adapting sequence labeling methods to jointly extract sentiment expressions, targets, and holders. The first work addressing the full SSA setting was presented by \citet{barnes-etal-2018-multibooked}, where these elements were represented using a BIO encoding and extracted with a CRF-based sequence labeler, while polarity was predicted using a linear SVM. Later, \citet{ovrelid-etal-2020-fine} explored a lossy BIO encoding combined with a BiLSTM encoder and a CRF inference layer.
\citet{barnes-etal-2021-structured} proposed, for the first time, casting the task as a dependency graph parsing problem, which they then solved using a graph-based biaffine parser \cite{dozat-manning-2018-simpler}---a state-of-the-art model for graph prediction in syntactic and semantic parsing---obtaining strong improvements over previous paradigms.
SSA was later introduced as a SemEval shared task~\cite{barnes-etal-2022-semeval}, attracting significant community interest, with 32 participating teams in total.
 
Among the best-performing systems, \citet{lin-etal-2022-zhixiaobao} adopted the graph-based baseline from \citet{barnes-etal-2021-structured}, replacing the BiLSTM encoder with pretrained MLMs (RoBERTa and XLM-RoBERTa for the English and non-English datasets, respectively). They further introduced an attention mechanism to handle tokens not included in any sentiment span and suffix masking to address split tokens. Other strong approaches, such as \citet{chen-etal-2022-mt}, also relied on the biaffine parser with XLM-RoBERTa, while incorporating several data augmentation strategies, including language modeling-based augmentation and the combination of Portuguese and English SemEval Laptop datasets \cite{pontiki-etal-2014-semeval}. Similarly, \citet{morio-etal-2022-hitachi} explored ensemble methods to compare graph-based and sequence-to-sequence approaches. Differently, \citet{barikbin-2022-slpl} proposed a partial sequence labeling approach using RoBERTa for English datasets and LaBSE for non-English datasets. Unlike graph-based parsers, this method follows a modular pipeline that uses sequence labeling to extract entity spans and attention-based scoring for relation extraction.

More recently, in subsequent work beyond the shared task, \citet{li2025extracting} extracted sentiment quadruples (holder, target, opinion, polarity) via a boundary‑sensitive token‑pair labeling framework that classifies head–tail, head–head, and tail–tail relations. Their system integrates syntax via conditional layer normalization, jointly labels polarity in the same space, and avoids graph parsing.

In this work, we follow previous research that formulates SSA as a dependency graph parsing task \cite{barnes-etal-2021-structured, barnes-etal-2022-semeval}. However, existing approaches rely on specialized graph parsers with task-specific attention or decoding mechanisms. In contrast, we investigate, for the first time, whether sequence labeling can provide a conceptually simpler and more flexible alternative for predicting target dependencies by relying solely on masked language models and simple feed-forward output layers.\footnote{While recent advances in large language models have shown promising capabilities in structured prediction tasks, this work focuses on supervised approaches that explicitly model the task formulation and exploit annotated SSA resources. This choice enables direct comparison with previous graph-based methods and provides a controlled framework to investigate alternative representations of sentiment dependencies, avoiding potential data contamination and memorization issues.} Building on recently proposed graph linearization strategies \cite{ezquerro-etal-2024-dependency}, we design an end-to-end encoder-decoder architecture that combines the representational power of MLM-based encoders with the simplicity and efficiency of sequence labeling decoders. This reformulation casts SSA as a token-level classification problem, eliminating the need for complex graph parsing components while preserving the ability to capture dependency structures.

\section{Dependency graph parsing as sequence labeling}\label{subsec:graph-linearizations}

In this work, we build on sentiment dependency graph representations and explore sequence labeling as an alternative to graph-based parsing. This allows us to perform SSA using only n tagging actions while relying on a simpler decoding process. Our approach starts from the standard dependency graph formulation of SSA and encodes the graph as a sequence of labels whose length matches the number of nodes. 

Formally, a dependency graph is defined as $G=(W,A)$, where $W=(w_1,...,w_n) \in\mathcal{V}^n$ is the input sequence of tokens
and $A$ is the set of arcs representing some type of semantic relation between the nodes of $W$. Each arc is denoted as $(h\overset{r}{\to}d)$, where $h\in[0,n]$ is the head or parent, $d\in[1,n]$ is the dependent (such that $h\neq d$) and $r\in\mathcal{R}$\footnote{We use $\mathcal{R}$ to denote the set of possible relations.} is the relation type between $w_h$ and $w_d$. 

Recently, \citet{ezquerro-etal-2024-dependency} have proposed several graph linearizations to transform the information of $A$ into a sequence of $n$ labels, denoted as $\ell=(\ell_1,...,\ell_n)\in\mathcal{L}^n$, and perform the reverse operation to recover $A$ from $\ell$.\footnote{We use $\mathcal{L}$ to denote the label set.} Formally, the \textit{encoding} process is an injective function\footnote{We use $\mathcal{A}^n$ to denote the set of arc sets for a sequence of $n$ nodes.} $\varepsilon:\mathcal{A}^n\to\mathcal{L}^n$ that transforms any set of arcs $A$ into a sequence of labels defined over $\mathcal{L}$, thus facilitating the design of a parameterized model that learns how to map $\mathcal{V}^n$ to $\mathcal{L}^n$ with linear complexity. The \textit{decoding} process instead is a 
function $\delta:\mathcal{L}^n\to\mathcal{A}^n$ that recovers the original set of arcs from the encoded labels, so $A=\delta(\varepsilon(A))$. 
These graph linearization strategies proposed by \citeauthor{ezquerro-etal-2024-dependency} can be broadly divided into two categories: \textit{unbounded} and \textit{bounded} encodings. Unbounded encodings represent graph structures using labels whose space grows with the size or density of the input graph, whereas bounded encodings constrain the representation to a fixed label space regardless of graph size. The main difference between these two families lies in the trade-off between representational flexibility and a controlled output space.
Next, we describe the encoding and decoding procedures for each encoding family. 

\subsection{Unbounded encodings}
These linerarizations propose to encode directly head positions or neighboring arcs to represent graphs as labels.

\begin{figure}
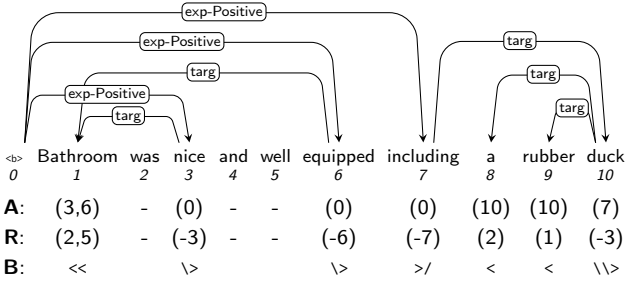
 \footnotesize
	\centering
	\resizebox{\columnwidth}{!}{%
		\hspace{-0.9em}
        \begin{dependency}
            \begin{deptext}[column sep=1pt]
                {\tiny\texttt{<b>}} \& {\scriptsize Bathroom} \& {\scriptsize was} \& {\scriptsize nice} \& {\scriptsize and} \& {\scriptsize well} \& {\scriptsize equipped} \& {\scriptsize including} \& {\scriptsize a} \& {\scriptsize rubber} \& {\scriptsize duck} \\[0.2em]
                \textsuperscript{\textit{0}}\& \textsuperscript{\textit{1}} \& \textsuperscript{\textit{2}} \& \textsuperscript{\textit{3}} \& \textsuperscript{\textit{4}} \& \textsuperscript{\textit{5}}\& \textsuperscript{\textit{6}} \& \textsuperscript{\textit{7}} \& \textsuperscript{\textit{8}} \& \textsuperscript{\textit{9}} \& \textsuperscript{\textit{10}}\\[0.3em]
                \textbf{A}: \& (3,6) \& - \&  (0) \& - \& - \& (0) \& (0) \& (10) \& (10) \& (7)  \\[0.4em]
                \textbf{R}: \& (2,5) \& - \& (-3) \& - \& - \& (-6) \& (-7) \& (2) \& (1) \& (-3)\\[0.4em]
                \textbf{B}: \& \texttt{<{}<} \& \& \texttt{\symbol{92}>} \& \& \& \texttt{\symbol{92}>} \& \texttt{>/} \& \texttt{<} \& \texttt{<} \& \texttt{\symbol{92}\symbol{92}>} \\
            \end{deptext}
            \depedge[edge unit distance=1.8ex]{1}{7}{exp-Positive}
            \depedge[edge unit distance=2.0ex]{1}{8}{exp-Positive}
            \depedge[edge unit distance=1.5ex]{7}{2}{targ}
            \depedge[edge unit distance=1.4ex]{4}{2}{targ}
            \depedge{11}{9}{targ}
            \depedge{11}{10}{targ}
            \depedge{8}{11}{targ}
            \depedge[edge unit distance=1.7ex, label style={fill=white}]{1}{4}{exp-Positive}
        \end{dependency}%
	}
	\caption{Absolute (\textbf{A}), relative (\textbf{R}) and bracketing (\textbf{B}) encodings for an example extracted from \cite{barnes-etal-2022-semeval}. Note that the \texttt{<b>} token is needed to identify the roots of the graph.}
	\label{fig:unbounded-example}
\end{figure}

On the one hand, the \textbf{absolute} and \textbf{relative} indexing consider the sorted sequence of head positions of each node $w_i$.  Figure \ref{fig:unbounded-example} shows an example of the unbounded encodings: the node $w_1$ (\textit{Bathroom}) has two parents: $w_3$ (\textit{nice}) and $w_6$ (\textit{equipped}), so its absolute label is $\ell_1=(3,6)$, where positions are always sorted to avoid redundancy of different arrangements of the same set of positions. The relative label subtracts to each index the actual position of the dependent, so $\ell_1=(2,5)$.

On the other hand, the \textbf{bracketing} encoding represents with different symbols the presence of an incoming or outgoing arc in some direction for each node. Formally, given the set of symbols $B=\{\texttt{\symbol{92}}, \texttt{>}, \texttt{<}, \texttt{/}\}$, the presence of \texttt{\symbol{92}} in $\ell_i$ indicates an outgoing arc from $w_i$ to the left, $\texttt{>}$ for an incoming arc from the left to $w_i$, \texttt{<} for an incoming arc from the right to $w_i$ and \texttt{/} for an outgoing arc from $w_i$ to the right. Figure \ref{fig:unbounded-example} shows an example of this encoding: $w_1$ has two parents located at the right, so two left-arrow symbols (\verb@<<@) are used for $\ell_1$. The parents of $w_1$ are obtained by looking at the following labels: $\ell_3=\texttt{\symbol{92}>}$, indicating that there is an outgoing arc to the left, so $w_3$ is parent of previous nodes whose labels contain the symbol \texttt{<}, thus resolving that $(3\to 1)\in A$.

\begin{figure}
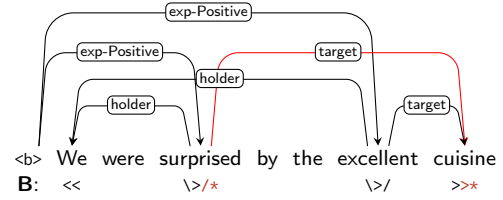
\footnotesize\centering 
	\centering
		\begin{dependency}
            \begin{deptext}[column sep=2pt]
                \texttt{<b>} \& We \& were \& surprised \& by \& the \& excellent \& cuisine \\[0.2em]
                \textbf{B}: \& \texttt{<{}<}  \& \& \texttt{\symbol{92}>\textcolor{BrickRed}{/*}} \& \& \&  \texttt{\symbol{92}>/} \& \texttt{>\textcolor{BrickRed}{>*}}\\
            \end{deptext}
            \depedge[edge height=35pt]{1}{4}{exp-Positive}
            \depedge[edge height=50pt]{1}{7}{exp-Positive}
            \depedge[edge height=15pt]{4}{2}{holder}
            \depedge[edge height=35pt, edge style={red}]{4}{8}{target}
            \depedge[edge height=15pt]{7}{8}{target}
            \depedge[edge height=25pt]{7}{2}{holder}
        \end{dependency}
    \caption{Graph example with two crossing arcs in the same direction: $(0 \to 6)$ crosses  $(3\to 7)$. For the bracketing (\textbf{B}) encoding, $(3\to 7)$ is encoded with a different set of brackets (\textcolor{BrickRed}{\texttt{/*}}, \textcolor{BrickRed}{\texttt{>*}}). The $6k$ bit encodings require $k=2$, so each plane is encoded with an independent sequence of bits.}
	\label{fig:example-cross}
\end{figure}

\paragraph{Decoding} Recovering the original set of arcs from the sequence of labels is straightforward for the absolute and relative indexing, since we only need to consider the index of the label and the positions stored in it. The bracketing encoding instead requires a 
single linear left-to-right pass using two stacks
to parse the sequence of brackets, pushing opening left positions into the left stack  (those where symbol \texttt{<} is found, indicating the opening of a left arc) and opening right positions into the right stack (where symbol \texttt{/} is found to open a right arc), and removing them when matching brackets are found (\texttt{\symbol{92}} closes a left arc with the element at the top of the left stack and \texttt{>} closes a right arc with the element at the top of the right stack).  
The bracketing linearization does naturally not support crossing arcs in the same direction, since the closing bracket would be resolved with the opening bracket of the crossing arc. To solve this limitation,
$A$ can be distributed into mutually-exclusive subsets of non-crossing arcs in the same direction (planes) and then independently encoding each plane with different variants of the original set of brackets.\footnote{For instance, the second plane uses $B^*=\{\texttt{\symbol{92}}^*, \texttt{>}^*, \texttt{<}^*, \texttt{/}^*\}$.} An example of this is displayed in Figure~\ref{fig:example-cross}.

\subsection{Bounded encodings}
Despite their simplicity, one of the main limitations of positional and bracketing encodings is that the cardinality of $\mathcal{L}$ grows indefinitely with the size or density of the graph. 
To counteract this, the \textbf{\boldmath 4$k$-bit encoding} and the \textbf{\boldmath 6$k$-bit encoding} enable a bounded label space, where each label is represented as a sequence of $m$ bits—$4k$ and $6k$, respectively—limiting the size of $\mathcal{L}$ to $2^m$.

\begin{figure}
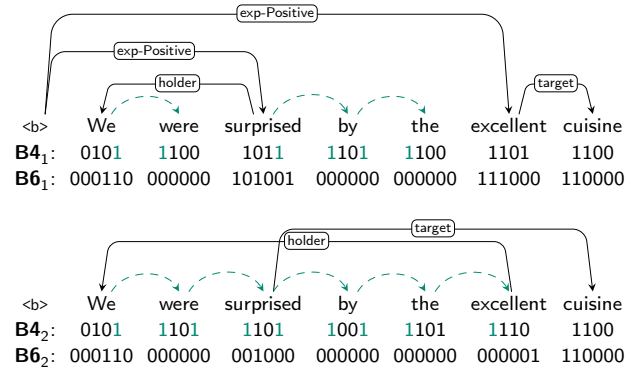
\centering\footnotesize 
    \resizebox{\columnwidth}{!}{%
    \begin{dependency}
        \begin{deptext}[column sep=2pt]
            \texttt{<b>} \& We \& were \& surprised \& by \& the \& excellent \& cuisine \\[0.3em]
            \textbf{B4}\textsubscript{1}: \& 010\textcolor{PineGreen}{1}  \& \textcolor{PineGreen}{1}100 \& 101\textcolor{PineGreen}{1}  \& \textcolor{PineGreen}{1}10\textcolor{PineGreen}{1} \& \textcolor{PineGreen}{1}100 \& 1101 \& 1100 \\[0.3em]
            \textbf{B6}\textsubscript{1}: \& 000110 \& 000000 \& 101001 \& 000000 \& 000000 \& 111000 \& 110000 \\
        \end{deptext}
        \depedge[edge height=25pt]{1}{4}{exp-Positive}
        \depedge[edge height=40pt]{1}{7}{exp-Positive}
        \depedge[edge height=12pt]{4}{2}{holder}
        \depedge[edge height=12pt]{7}{8}{target}

        \depedge[text only label, arc edge, edge style={PineGreen, dashed}]{2}{3}{}
        \depedge[text only label, arc edge, edge style={PineGreen, dashed}]{4}{5}{}
        \depedge[text only label, arc edge, edge style={PineGreen, dashed}]{5}{6}{}
    \end{dependency}%
    }
    \\[0.8em]
    \resizebox{\columnwidth}{!}{%
    \begin{dependency}
        \begin{deptext}[column sep=2pt]
            \texttt{<b>} \& We \& were \& surprised \& by \& the \& excellent \& cuisine \\[0.3em]
            \textbf{B4}\textsubscript{2}: \& 010\textcolor{PineGreen}{1} \& \textcolor{PineGreen}{1}10\textcolor{PineGreen}{1} \& \textcolor{PineGreen}{1}10\textcolor{PineGreen}{1} \& \textcolor{PineGreen}{1}00\textcolor{PineGreen}{1} \& \textcolor{PineGreen}{1}101 \& \textcolor{PineGreen}{1}110 \& 1100 \\[0.3em]
            \textbf{B6}\textsubscript{2}: \& 000110 \& 000000 \& 001000 \& 000000 \& 000000 \& 000001 \&  110000 \\
        \end{deptext}
        \depedge[edge height=25pt]{4}{8}{target}
        \depedge[edge height=20pt, edge horizontal padding=-1pt]{7}{2}{holder}

        \depedge[text only label, arc edge, edge style={PineGreen,dashed}]{2}{3}{}
        \depedge[text only label, arc edge, edge style={PineGreen,dashed}]{3}{4}{}
        \depedge[text only label, arc edge, edge style={PineGreen,dashed}]{4}{5}{}
        \depedge[text only label, arc edge, edge style={PineGreen,dashed}]{5}{6}{}
        \depedge[text only label, arc edge, edge style={PineGreen,dashed}]{6}{7}{}
    \end{dependency}%
    }
    \caption{$4k$ (\textbf{B4}\textsubscript{$k$}) and $6k$-bit (\textbf{B6}\textsubscript{$k$}) encoding example from the graph displayed in Figure \ref{fig:example-cross}. Note that the arcs of the original graph need to be distributed in two different subsets that are independently encoded. Artificial arcs (green dashed lines) are generated in the $4k$-bit encoding to satisfy the first assumption of the algorithm (each node needs one and only one parent). Those arcs are ignored for \textbf{B6}\textsubscript{1} and \textbf{B6}\textsubscript{2}. Note that in this case the subsets of the $4k$ and $6k$-bit encoding are the same, but this might not happen for other graphs.}
    \label{fig:example-bit}
\end{figure}

Both encodings assume certain constraints on the arcs of a dependency graph $G$ to enable correct encoding and decoding using 4 and 6 bits per label, respectively. The $4k$-bit encoding covers dependency graphs where (i) each node has exactly one parent and (ii) no arc crosses another arc in the same direction. The $6k$-bit encoding, in contrast, supports dependency graphs where (i) each node has at most one parent per direction and (ii) no arc crosses another arc in the same direction.
Since these constraints considerably limit graph expressiveness, the arcs of $A$ can be distributed into at most $k$ subsets, denoted as ${P_1\dots P_k}$, each satisfying the corresponding constraints.
Each subset is then independently encoded with 4 (or 6) bits per label and the resulting $k$ sequences are concatenated at the token level to produce $n$ labels of $4k$ and $6k$ bits each (see example in Figure \ref{fig:example-bit}). Our preliminary experiments indicated that the 4k-bit encoding underperformed the baselines, likely due to the overhead of creating numerous artificial arcs. Consequently, we selected the 6k-bit version as the representative for this family of bounded encodings.

In 6$k$-bit, the arcs of $P_j$ that are associated with the node $w_i$, for $i=1\dots n$ and  $j=1\dots k$, determine the bit values of the $i$th label in the $j$th label sequence. 
The $6k$-bit encoding independently encodes each subset $P_j$ with 6 bits per label. The $i$th label associated to $P_j$ is denoted as 
$\ell_i^j=(b_i^0, b_i^1, b_i^2, b_i^3, b_i^4, b_i^5)$, where $b_i^0$ ($b_i^3$) is activated if $w_i$ has a left (right) parent, $b_i^1$ ($b_i^4$) if $w_i$ is the outermost dependent of its right (left) parent, and $b_i^2$ ($b_i^5$) if $w_i$ has right (left) dependents. As the $6k$-bit encoding requires to first distribute the arcs of $A$ into $k$ subsets, and then independently encode each subset considering the definition of $\ell_i^j$. Figure \ref{fig:example-bit} shows in B6\textsubscript{1} and B6\textsubscript{2} the $6k$-bit encoding with $k=2$.\footnote{Note that, unlike the $4k$-bit encoding, there are no restrictions on the minimum number of parents per node, so the $6k$-bit encoding does not generate artificial arcs.}
 
In 6$k$-bit encoding, the parameter $k$ can be adjusted to find a good tradeoff between compactness (size of $\mathcal{L}$) and coverage: a smaller $k$ yields a smaller label set, hence improving efficiency and learnability, but may not be able to represent dense graphs or those with complex patterns of mutually crossing arcs, due to impossibility of splitting them into $k$ subsets meeting the restrictions. However, \citet{ezquerro-etal-2024-dependency} empirically showed that $k=2$ or $3$ is usually sufficient to provide a very high coverage on the datasets they tested.

\section{System overview}

\begin{figure*}[tbp!]\centering
    \includegraphics[width=\linewidth]{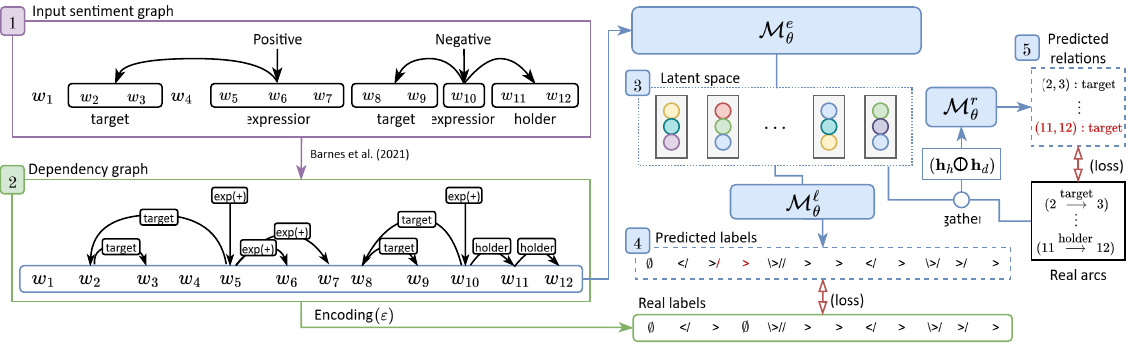}
    \caption{\label{fig:pipeline-train}Visualization of the system at training time. Each module is represented with a different color: purple for the sentiment-to-dependency transformation, green for the graph linearization and blue for the neural network. Hypothetical errors are displayed in red for completeness. The input is a sentiment graph (\purplebox{1}) that is converted into a dependency graph (\greenbox{2}) through \citet{barnes-etal-2021-structured}. The input sentence is fed to $\mathcal{M}_\theta^e$  and $\mathcal{M}_\theta^\ell$ to obtain the hidden embeddings (\bluebox{3}) and the predicted labels (\bluebox{4}). The arcs from \greenbox{2} are used to pair embeddings and feeding them to $\mathcal{M}_\theta^r$ (\bluebox{5}). The network is optimized with the cross-entropy loss of the predicted labels and predicted relations.}
\end{figure*}

\begin{figure*}[tbp!]\centering
    \includegraphics[width=\linewidth]{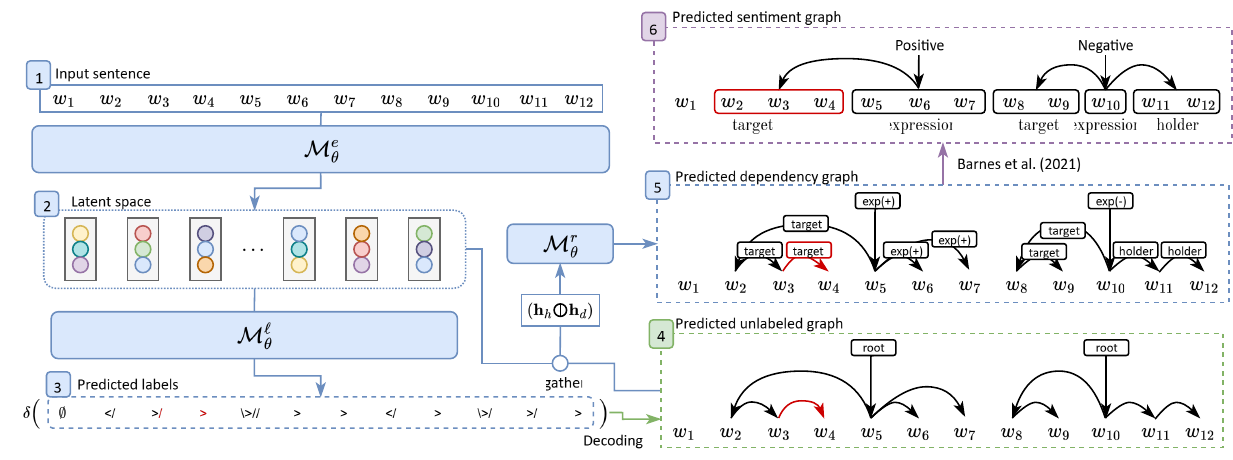}
    \caption{\label{fig:pipeline-inference}Visualization of the system at inference time. Same color notation as in Figure \ref{fig:pipeline-train}. The input is a sentence (\bluebox{1}). The embeddings and predicted labels are obtained by the trained system, as depicted in Figure \ref{fig:pipeline-train}. The decoding process is applied to the predicted labels to recover an unlabeled dependency graph. The $r$ component of each predicted arc is obtained by gathering the corresponding embeddings of \bluebox{2} and feed them to $\mathcal{M}_\theta^r$. The predicted dependency graph (\bluebox{5}) is transformed into a sentiment graph (\purplebox{6}) through the inverse transformation of \cite{barnes-etal-2021-structured}.}
\end{figure*}

Our system relies on two phases to perform SSA through dependency graph parsing formulated as a sequence labeling task: (i) the sentiment-to-dependency transformation, which converts sentiment structures into dependency graphs, and (ii) the dependency graph linearization and sequence labeling model, which encodes the graph structure as token-level labels and learns to predict them with a neural network. In this section, we detail each module and how they are used to learn the sentiment graph structure from an input sentence (Figures~\ref{fig:pipeline-train} and~\ref{fig:pipeline-inference}).

\subsection{Sentiment-to-dependency transformation} Let $\mathcal{G}=(W,\Omega)$ be an input sentiment graph for the sentence $W=(w_1,..,w_n)\in\mathcal{V}^n$, where $\Omega=(O_1,...,O_m)$ is the set of opinion tuples, as described in Section \ref{introduction}. \citet{barnes-etal-2021-structured} proposed two strategies to transform each opinion tuple $O'=(h,t,e,p)\in\Omega$ into a valid set of dependency arcs $A'$ and recover $O'$ from $A'$. The complete set $A$ is the union of  subsets $A'$ obtained after transforming all opinion tuples of $\Omega$.

Let $x_1$ and $x_{-1}$, where $x\in\{h,t,e\}$ be the first and last token, respectively, of the holder, target or expression components. The first approach, named \textbf{head-first}, uses  $x_1$ as the head of the other tokens in $x$, and then connects the three components with $\{0\overset{p}{\to} e_1, e_1\overset{\text{holder}}{\to} h_1, e_1\overset{\text{target}}{\to} t_1\}$. See an example in Figure \ref{fig:ssa-graph}: the expression ``\textit{in great condition}'' has \textit{in} as the first token, so  ``\textit{in}'' is connected to ``\textit{great}'' and ``\textit{condition}'' as the parent node. Thus, for the opinion tuple $O_1=(\lambda,\textit{the hotel}, \textit{in great condition}, +)$, the target is represented with the arcs $\{\textit{the}\overset{\text{target}}{\longrightarrow} \textit{hotel}\}$, the expression with $\{\textit{in}\overset{\text{exp(+)}}{\longrightarrow}\textit{great}, \textit{in}\overset{\text{exp(+)}}{\longrightarrow} \textit{condition}\}$ and both are connected with $\{0\overset{\text{exp(+)}}{\longrightarrow}\textit{in},\textit{in}\overset{\text{target}}{\longrightarrow}\textit{the}\}$, . In the \textbf{head-final} approach, instead of using the first token $x_1$ to connect different components, it uses the last token, so the same opinion tuple is represented as $A'=\{\textit{hotel}\overset{\text{target}}{\longrightarrow}\textit{the}, \textit{condition}\overset{\text{exp(+)}}{\longrightarrow}\textit{in}, \textit{condition}\overset{\text{exp(+)}}{\longrightarrow}\textit{great}, 0\overset{\text{exp(+)}}{\longrightarrow}\textit{condition},\textit{condition}\overset{\text{target}}{\longrightarrow}\textit{hotel}\}$. 

It is possible to recover the original opinion tuple $O'$ from $A'$, since it only requires first identifying the holder, target and expression as those dependents with an arc relation of each type and then use the arc relation of the root node to extract the polarity. The conversion becomes lossy when multiple opinion tuples span through common tokens, creating multiple re-entrancies for the same token. In these cases, the recovering algorithm is not able to identify  which arc needs to be considered to reconstruct each opinion tuple. This phenomenon is rare in the datasets considered in this paper, affecting only a small proportion of graphs. Importantly, this limitation is inherent to representing SSA as a dependency graph rather than to our parsing approach: whenever multiple opinion tuples induce indistinguishable graph structures, the original tuples cannot be uniquely recovered from the graph alone. Consequently, all approaches that formulate SSA as a graph parsing problem, regardless of the parsing architecture or algorithm employed, exhibit the same out-of-coverage cases. Exact coverage statistics are provided later in the paper (see the last column of Table~\ref{tab:coverage}).

\subsection{Dependency graph linearization and neural tagger} Given the dependency graph $G=(W,A)$ obtained from the previous step, the information of $A$ is encoded as $\varepsilon(A)=(\ell_1,...,\ell_n)\in\mathcal{L}^n$ through a graph linearizations introduced in Section \ref{subsec:graph-linearizations}.  Thus, to predict $A$, our system uses a neural network to learn the distribution of $\mathcal{L}^n$ from $\mathcal{V}^n$. We followed the encoder-decoder neural framework, where the encoder $\mathcal{M}_\theta^e:\mathcal{V}^n\to\mathbb{R}^{n\times d}$,\footnote{We use the symbol $\theta$ to denote the set of learnable parameters of a neural network.} learns a $d$-dimensional contextualized projection of each token, and the decoder $\mathcal{M}^\ell_\theta:\mathbb{R}^{d}\to\mathcal{L}$ independently processes each representation to predict a label.  

Note that in the linearizations introduced in Section \ref{subsec:graph-linearizations}, $\delta$ is not able to recover the $r$ component in each arc $(h \overset{r}{\to} d)$. To effectively learn the probability distribution over $\mathcal{R}$, we added an additional module, $\mathcal{M}_\theta^r: (\mathbb{R}^d, \mathbb{R}^d) \to \mathcal{R}$, which concatenates the hidden representations of two words, $(w_h, w_d)$, and predicts their relationship type, provided that an arc $(h \overset{r}{\to} d)$ exists. 

Figures \ref{fig:pipeline-train} and \ref{fig:pipeline-inference} show a training and prediction example of our system, respectively. At training time (Figure \ref{fig:pipeline-train}),  
each training instance is a sentence $W$ together with its complete gold sentiment graph. This graph is converted through \citet{barnes-etal-2021-structured}'s transformation into a dependency graph $G=(W,A)$. In turn, said dependency graph is passed through the sequence-labeling encoding to obtain a sequence of gold labels $\varepsilon(A)$, which is the desired output we will train our neural network to produce. 

First, the sequence of tokens $W$ is fed to a neural encoder to recover the contextualized embeddings $(\mathbf{h}_1,...,\mathbf{h}_n)\in\mathbb{R}^{n\times d}$. Each embedding is fed to $\mathcal{M}_\theta^\ell$ to obtain a sequence of predicted labels $\hat{\ell}=(\hat{\ell}_1,..,\hat{\ell}_n)\in\mathcal{L}^n$; and the cross-entropy loss between $\varepsilon(A)$  and $\hat{\ell}$ is computed to optimize $\theta$. In order to guide $\mathcal{M}_\theta^r$ to learn $\mathcal{R}$ from two paired words, for each arc $(h\overset{r}{\to}d)\in A$, the system gathers and concatenates $(\mathbf{h}_h,\mathbf{h}_d)$ and feeds them to $\mathcal{M}_\theta^r$ to learn $\mathcal{R}$.

At inference time (Figure \ref{fig:pipeline-inference}), the input to our system is instead a sentence with no arc information. The embedding matrix and predicted labels are obtained with $\mathcal{M}_\theta^e$ and $\mathcal{M}_\theta^\ell$. Then, the system runs the decoding process using the sequence of predicted labels and, for each decoded arc, gathers the corresponding embeddings and obtains its relationship type $r$ with $\mathcal{M}_\theta^r$, thus obtaining a dependency graph. Finally, this predicted dependency graph is transformed into a sentiment graph through \citet{barnes-etal-2021-structured}'s inverse transformation.

\section{Experimental setup}
We next describe the datasets used, the required pre-processing, and the training setup.
\subsection{Datasets}

We conduct experiments on the five datasets provided in the SemEval-2022 Task 10: the OpeNER dataset \cite{agerri-etal-2013-opener}
with English and Spanish annotations, 
the English Multi-Perspective Question Answer corpus \cite[MPQA;][]{wiebe2005annotating} and the Darmstadt Universities corpus \cite[DS;][]{toprak-etal-2010-sentence}, the Norwegian NoReC dataset \cite{velldal-etal-2018-norec} and the Basque and Catalan MultiBooked dataset \cite[MB;][]{barnes-etal-2018-multibooked}.
The main statistics of these datasets are presented in the Table \ref{tab:dataset_stats}. 

Table~\ref{tab:coverage} reports the coverage of each encoding on each dataset, i.e., the percentage of dependency graphs that can be recovered with each encoding (or, equivalently, the maximum F1-score achievable if the sequence labeling system made no prediction errors). Coverage may be lower than 100\% for two reasons: (i) the sentiment-to-dependency transformation is not lossless (last column), or (ii) the sequence-labeling encodings cannot represent certain dense graphs or graphs with crossing arcs. Overall, the results show that setting $k=2$ or $3$, depending on the dataset, is sufficient to achieve coverage above 99\%. The only exceptions are MPQA\textsubscript{en} and NoReC\textsubscript{no}, where coverage is lower. However, in both datasets, most of the missing coverage is due to the sentiment-to-dependency transformation.

Finally, Table \ref{tab:label-space} presents the size of the label space for each encoding and representation; and number of unique arc labels.

\begin{table}[htb!]\centering\footnotesize
    \setlength{\tabcolsep}{3pt}
    \renewcommand{\arraystretch}{1.2}
    \caption{General statistics of the datasets used in our experimental study. Subscripts denote the language denoted by its ISO-639 code. Different splits in each subrow: train, development and test. Columns $n$,  $|O|$ and $|A|/n$ denote the average sentence length, number of opinion tuples and ratio of number of arcs and nodes, respectively.} \label{tab:dataset_stats}
    \begin{tabular}{|p{0.2cm}|p{0.05cm}|cccc|ccc|}
        \cline{3-9}
        \multicolumn{2}{c|}{}&\textbf{\#sents} & \textbf{\#holders} & \textbf{\#targets} & \textbf{\#expr} & $n$ &  $|O|$ & $|A|/n$ \\
        \hline
        \multicolumn{2}{|c|}{\parbox[t]{2mm}{\multirow{3}{*}{\rotatebox[origin=c]{90}{DS\textsubscript{en}}}}}
        & 2253 & 69 & 869 & 1525 & 19.99 & 0.36 & 0.07 \\
        \multicolumn{2}{|c|}{}& 232 & 8 & 117 & 186 & 18.09 & 0.42 & 0.09 \\
        \multicolumn{2}{|c|}{}& 318 & 12 & 155 & 251 & 20.40 & 0.41 & 0.08 \\
        \hline 
        \multicolumn{2}{|c|}{\parbox[t]{2mm}{\multirow{3}{*}{\rotatebox[origin=c]{90}{MPQA\textsubscript{en}}}}}
        & 5581 & 2925 & 8625 & 3147 & 24.37 & 0.28 & 0.10 \\
        \multicolumn{2}{|c|}{}& 1979 & 852 & 2499 & 1024 & 23.91 & 0.27 & 0.08 \\
        \multicolumn{2}{|c|}{} & 2025 & 894 & 2696 & 958 & 23.85 & 0.24 & 0.08 \\
        \hline 
        \parbox[t]{2mm}{\multirow{6}{*}{\rotatebox[origin=c]{90}{MB}}} & \parbox[t]{2mm}{\multirow{3}{*}{\rotatebox[origin=c]{90}{ca}}} 
        & 1174 & 149 & 3355 & 4866 & 15.62 & 1.60 & 0.53 \\
        && 167 & 20 & 412 & 626 & 13.37 & 1.45 & 0.52 \\
        && 335 & 43 & 912 & 1348 & 14.78 & 1.47 & 0.53 \\
        \cline{2-9}
        & \parbox[t]{2mm}{\multirow{3}{*}{\rotatebox[origin=c]{90}{eu}}}
        & 1063 & 180 & 1349 & 3586 & 10.52 & 1.50 & 0.53 \\
        && 152 & 28 & 149 & 492 & 10.70 & 1.30 & 0.50 \\
        && 305 & 54 & 347 & 934 & 10.73 & 1.35 & 0.48 \\
        \hline 
        \multicolumn{2}{|c|}{\parbox[t]{2mm}{\multirow{3}{*}{\rotatebox[origin=c]{90}{NRC\textsubscript{no}}}}}
        & 8634 & 632 & 9670 & 37300 & 16.71 & 0.90 & 0.32 \\
        \multicolumn{2}{|c|}{} & 1531 & 86 & 1652 & 6724 & 16.92 & 0.88 & 0.33 \\
        \multicolumn{2}{|c|}{} & 1272 & 67 & 1441 & 5504 & 17.18 & 0.91 & 0.32 \\
        \hline 
        \parbox[t]{2mm}{\multirow{6}{*}{\rotatebox[origin=c]{90}{OpeNER}}} & \parbox[t]{2mm}{\multirow{3}{*}{\rotatebox[origin=c]{90}{en}}} 
        & 1744 & 203 & 3866 & 6359 & 14.72 & 1.53 & 0.48 \\
        && 249 & 40 & 547 & 911 & 14.22 & 1.54 & 0.49 \\
        && 499 & 77 & 1158 & 1907 & 15.49 & 1.60 & 0.49 \\
        \cline{2-9}
        & \parbox[t]{2mm}{\multirow{3}{*}{\rotatebox[origin=c]{90}{es}}}
        & 1438 & 118 & 4990 & 5940 & 17.13 & 1.89 & 0.54 \\
        && 206 & 18 & 734 & 797 & 17.08 & 1.70 & 0.50 \\
        && 410 & 42 & 1559 & 1912 & 18.31 & 2.12 & 0.55 \\
        \hline
    \end{tabular}
\end{table}

\begin{table}[h!]\centering\footnotesize
    \caption{Coverage (SF\textsubscript{1}) of each encoding in the test split. $\Omega$ represents the SF\textsubscript{1} of the original sentiment to dependency transformation. The first and second subrows for each dataset are for the head-first and head-final approach, respectively.}\label{tab:coverage}
    \resizebox{\columnwidth}{!}{%
    \setlength{\tabcolsep}{3pt}
    \renewcommand{\arraystretch}{1.1}
    \begin{tabular}{|r|ccc|ccc|c|}
        \cline{2-8}
        \multicolumn{1}{c|}{}& \multicolumn{3}{c|}{\textbf{Br.}} & \multicolumn{3}{c|}{\textbf{$6k$-bit}} & \multirow{2}{*}{$\Omega$} \\ 
        \cline{2-7}
        \multicolumn{1}{c|}{}& 1 & 2 & 3 & 1 & 2 & 3 & \\
        \hline 
        DS\textsubscript{en} 
            & 99.0 & 99.8 & 99.8 & 92.9 & 99.8 & 99.8 & 99.8 \\ 
            & 99.0 & 99.8 & 99.8 & 92.5 & 99.8 & 99.8 & 99.8 \\
        MPQA\textsubscript{en} 
            & 92.4 & 95.5 & 95.9 & 83.1 & 94.2 & 95.7 & 95.9 \\ 
            & 90.4 & 96.5 & 96.5 & 82.3 & 95.0 & 96.5 & 96.5\\
        MB\textsubscript{ca} 
            & 99.1 & 99.8 & 99.8 & 86.1 & 97.9 & 99.4 & 99.8 \\
            & 98.6 & 99.8 & 99.8 & 85.9 & 97.9 & 99.4 & 99.8  \\
        MB\textsubscript{eu} 
            & 97.7 & 100 & 100 & 83.1 & 95.7 & 99.1 & 100 \\
            & 97.3 & 100 & 100 & 82.4 & 95.7 & 99.1 & 100 \\
        NoReC\textsubscript{no} 
            & 89.0 & 94.2 & 94.5 & 77.5 & 90.9 & 93.8 & 95.2\\
            & 87.5 & 93.0 & 93.0 & 73.0 & 88.9 & 92.2 & 93.6 \\ 
        OpeNER\textsubscript{en}  
            & 95.9 & 100 & 100 & 84.4 & 98.3 & 100  & 100\\
            & 95.9 & 100 & 100 & 84.4 & 98.3 & 99.9 & 100\\
        OpeNER\textsubscript{es} 
            & 99.1 & 100 & 100 & 88.4 & 98.4 & 100 & 100 \\
            & 98.9 & 100 & 100 & 88.1 & 98.3 & 100 & 100 \\
        \hline 
    \end{tabular}%
    }
\end{table}
\begin{table}[h!]\centering\footnotesize
    \caption{Size of the label space ($|\mathcal{L}|$) for each encoding and representation; and number of unique arc labels (\textbf{\#rels}). Same notation as in Table \ref{tab:results}.}\label{tab:label-space}
    \resizebox{\columnwidth}{!}{%
    \setlength{\tabcolsep}{2pt}
    \renewcommand{\arraystretch}{1.1}
    \begin{tabular}{|c|r|cc|ccc|ccc|c|}
        \cline{3-11}
        \multicolumn{2}{c|}{} & \multirow{2}{*}{\textbf{A}} & \multirow{2}{*}{\textbf{R}}& \multicolumn{3}{c|}{\textbf{Br.}} & \multicolumn{3}{c|}{\textbf{$6k$-bit}}& \multirow{2}{*}{\textbf{\#rels}} \\
        \cline{5-10}
        \multicolumn{2}{c|}{}& & & 1 & 2 & 3 & 1 & 2 & 3 & \\
        \hline 
        \parbox[t]{2mm}{\multirow{7}{*}{\rotatebox[origin=c]{90}{\textit{head-first}}}} 
        & DS\textsubscript{en} & 117 & 134 & 37 & 45 & 45 & 15 & 30 & 37 & 6 \\
        & MPQA\textsubscript{en} &  528 & 615 & 170 & 380 & 403 & 22 & 96 & 140 & 13 \\
        & MB\textsubscript{ca} & 303 & 274 & 96 & 138 & 138 & 19 & 55 & 72 & 4 \\
        & MB\textsubscript{eu} &  197 & 193 & 59 & 91 & 91 & 18 & 48 & 61 & 4 \\
        & NoReC\textsubscript{no} & 1047 & 1371 & 276 & 722 & 778 & 24 & 147 & 243 & 14 \\
        & OpeNER\textsubscript{en} & 355 & 338 & 83 & 136 & 140 & 18 & 57 & 80 & 4 \\
        & OpeNER\textsubscript{es} & 388 & 343 & 82 & 129 & 131 & 18 & 51 & 64 & 4 \\
        
        \hline 
        \parbox[t]{2mm}{\multirow{7}{*}{\rotatebox[origin=c]{90}{\textit{head-final}}}} 
        & DS\textsubscript{en} & 121 & 137 & 32 & 42 & 42 & 16 & 32 & 38 & 5 \\
        & MPQA\textsubscript{en} & 516 & 634 & 169 & 337 & 344 & 29 & 102 & 138 & 13 \\
        & MB\textsubscript{ca} & 316 & 288 & 81 & 119 & 119 & 17 & 45 & 59 & 4 \\
        & MB\textsubscript{eu} & 216 & 216 & 56 & 85 & 85 & 18 & 47 & 61 & 4 \\
        & NoReC\textsubscript{no} & 1222 & 1587 & 256 & 629 & 691 & 27 & 142 & 251 & 16 \\
        & OpeNER\textsubscript{en} & 411 & 375 & 79 & 123 & 127 & 18 & 48 & 74 & 4 \\
        & OpeNER\textsubscript{es} & 412 & 365 & 88 & 130 & 132 & 16 & 44 & 56 & 4 \\
        \hline 
    \end{tabular}%
    }
\end{table}


\subsection{Data preprocessing} To convert the sentiment graphs to dependency graphs, we used the official code\footnote{\url{https://github.com/jerbarnes/semeval22_structured_sentiment}} from the SemEval-2022 Task 10, testing both the head-first and head-final representations. As described above, in the head-first representation, the first token of the sentiment expression is treated as the root node, with the first tokens of the holder and target spans attached as its dependents. Similarly, in the head-final representation, the last token of the sentiment expression serves as the root node, and the last tokens of the holder and target spans are attached as dependents.

\subsection{Training setup and hyperparameters:} 
To evaluate the proposed sequence-labeling approach, we fine-tune a well-known multilingual pretrained encoder, which is a de facto model in many applications, as the underlying model for predicting the labels associated with each token. Specifically, we use the XLM-RoBERTa-large encoder \cite{conneau2020unsupervised}, a robust multilingual encoder pretrained on text in 100 languages. We fine-tune the model using a linear learning-rate schedule with an initial learning rate of $1\mathrm{e}{-5}$ and a batch size of 64, employing gradient accumulation over 10 steps. Training is performed with early stopping (patience of 20 epochs) and capped at a maximum of 50 epochs, using a convergence tolerance of 0.03 as the stopping criterion. The experiments are conducted on a single NVIDIA RTX 6000 Ada GPU.

\subsection{Evaluation}
To evaluate our sequence-labeling approach and enable comparisons with prior state-of-the-art systems, we adopt the two main evaluation protocols used in the literature. We first follow the official SemEval-2022 evaluation protocol, which enables direct comparison with systems participating in the shared task, and then report a more fine-grained set of metrics following prior work.

\paragraph{Protocol 1: Sentiment Graph F\textsubscript{1} Evaluation}
This protocol follows the official SemEval-2022 evaluation proposed by \citet{barnes-etal-2022-semeval}, reporting the labeled sentiment graph F\textsubscript{1} score (SF\textsubscript{1}) as the primary metric. It enables direct comparison (\S~\ref{sf1-evaluation}) with top-ranked systems from the shared task.

\paragraph{Protocol 2: Fine-grained metrics}
This protocol provides a more detailed view of model performance following \citet{barnes-etal-2021-structured}. In particular, we report: (i) token-level F\textsubscript{1} scores for holder, target, and expression spans; (ii) targeted F\textsubscript{1} (target expression with polarity); (iii) unlabeled and labeled dependency parsing F\textsubscript{1}; and (iv) sentiment F\textsubscript{1} with and without polarity (NSF\textsubscript{1} and SF\textsubscript{1}). These metrics enable us to better isolate where gains or losses originate, whether in span detection, relation labeling, or polarity assignment.

\paragraph{Internal baseline: Biaffine decoder}
To isolate the effect of our sequence-labeling formulation from other architectural choices, we include an internal baseline that replaces the sequence-labeling decoder with the standard biaffine decoder \citep{dozat-manning-2018-simpler} used by prior graph-based methods \citep{samuel-etal-2022-direct, lin-etal-2022-zhixiaobao, alonso-alonso-etal-2022-lys}, while keeping the same XLM-RoBERTa-large encoder. This allows us to attribute performance differences to the prediction formulation rather than to the encoder or training setup.

\section{Results}\label{Results}

We next analyze the results using both evaluation protocols, reporting both the official shared-task ranking metrics and the more fine-grained evaluation metrics used in related work predating the shared task.

\subsection{Labeled sentiment graph F\textsubscript{1} score evaluation} \label{sf1-evaluation}

Table~\ref{tab:results} presents the performance of our models on each dataset according to the official SemEval shared-task ranking metric (SF\textsubscript{1}) for both the \textit{head-first} (\textit{h.first}) and \textit{head-final} (\textit{h.final}) transformations.

\paragraph{Sequence labeling vs graph-based decoder}Compared with the biaffine graph-based baseline, our models outperform the graph-based approach on most treebanks under the \textit{head-first} strategy. The trend differs for the \textit{head-final} representation, which also appears to be better suited to biaffine parsers, yielding an average SF\textsubscript{1} of 57.8 compared with 51.8 under \textit{head-first}.

\begin{table*}[ht!]\centering\footnotesize
    \caption{SF\textsubscript{1} score in the test set for the absolute (\textbf{A}), relative (\textbf{R}), bracketing (\textbf{Br}) and \textbf{\boldmath $6k$-bit} encodings. Subcolumns in the bracketing and bit-based encodings denote the value of the hyperparameter $k$. The best result is highlighted in bold. The average is included in the last row ($\mu$).} \label{tab:results}
    \resizebox{\textwidth}{!}{
    \renewcommand{\arraystretch}{1.3}
    \setlength{\tabcolsep}{2pt}
    \begin{tabular}{|r|cc|ccc|ccc|c|cc|ccc|ccc|c|}
        \cline{2-19}
        \multicolumn{1}{c|}{}& \multicolumn{9}{c|}{\textit{h.first}} & \multicolumn{9}{c|}{\textit{h.final}} \\
        \cline{2-19}
        \multicolumn{1}{c|}{} & \multirow{2}{*}{\textbf{A}} & \multirow{2}{*}{\textbf{R}}& \multicolumn{3}{c|}{\textbf{Br.}} & \multicolumn{3}{c|}{\textbf{$6k$-bit}}& \multirow{2}{*}{\textbf{Biaf.}} & \multirow{2}{*}{\textbf{A}} & \multirow{2}{*}{\textbf{R}}& \multicolumn{3}{c|}{\textbf{Br.}} & \multicolumn{3}{c|}{\textbf{$6k$-bit}}& \multirow{2}{*}{\textbf{Biaf.}} \\
        \cline{4-9}
        \cline{13-18}
        \multicolumn{1}{c|}{}& & & 1 & 2 & 3 & 1 & 2 & 3 & & & & 1 & 2 & 3 & 1 & 2 & 3 & \\
        \hline 
        \textbf{DS}\textsubscript{en} & 37.9 & 19.8 & 44.0  & 41.4  & 41.4 & 43.8 & \textbf{46.1} & 43.3 & 35.0 & 42.1 & 18.6 & 36.5 & 39.8 & 40.9 & 41.2 & 38.9 & 41.2 & 44.3 \\
        \textbf{MPQA}\textsubscript{en} & 17.6 & 18.6 & 21.3  & 17.1  & 28.5 & \textbf{32.6} & 28.8 & 32.3 & 29.8 & 16.8 & 13.1 & 27.2 & 17.8 & 28.5 & 31.5 & 34.8 & 31.6 & 29.8 \\ 
        \textbf{MB}\textsubscript{ca} & 68.3 & 57.1 & 61.6  & 69.2  & 70.8 & 73.0 & \textbf{77.0} & 71.8 & 61.1 & 51.1 & 47.6 & 56.8 & 64.6 & 62.7 & 62.8 & 65.0 & 67.0 & 69.3 \\
        \textbf{MB}\textsubscript{eu} & 71.3 & 57.1 & 60.5  & 66.4 & 69.5 & \textbf{72.5} & 68.6 & 71.5 & 62.1 & 61.4 & 54.3 & 55.3 & 64.6 & 60.6 & 65.5 & 63.0 & 65.2 & 70.9 \\
        \textbf{NoReC}\textsubscript{no} & 48.9 & 38.5 & 40.9  & 45.2  & 42.4 & 47.3 & \textbf{51.9} & 49.5 & 48.0 & 42.8 & 31.1 & 39.1 & 46.3 & 48.7 & 50.1 & 49.0 & 49.7 & 47.1 \\
        \textbf{OpeNER}\textsubscript{en} & 72.8 & 61.7 & 66.6  & 69.3  & 68.2 & 74.0 & \textbf{75.7} & 73.4 & 65.8 & 59.8 & 53.1 & 63.3 & 67.2 & 66.4 & 68.9 & 70.0 & 71.7 & 72.8 \\ 
        \textbf{OpeNER}\textsubscript{es} & 72.5 & 55.8 & 62.1  & 71.9  & 68.0 & 72.7 & \textbf{74.8} & 72.9 & 60.7 & 51.7 & 46.6 & 61.9 & 65.5 & 65.5 & 63.0 & 66.7 & 66.5 & 70.5 \\
        \hline 
        $\mu$ & 55.6 & 44.1 &  51.0 & 54.4 & 55.5 & 59.4 & \textbf{60.4} & 59.2 & 51.8  & 46.5 & 37.8 & 48.6 & 52.3 & 53.3 & 54.7 & 55.3 & 56.1 & 57.8 \\
        \hline
    \end{tabular}
    }
\end{table*}

\begin{table}[ht!]\centering\footnotesize
    \caption{SF\textsubscript{1} comparison of different studies and our best-performing encoding ($6k$-bit with $k=2$). Systems proposed in the SemEval-2022 Task 10 \cite{barnes-etal-2022-semeval} are marked with an asterisk. Best results highlighted in bold.}\label{tab:results-comparison}
    \resizebox{\columnwidth}{!}{
    \renewcommand{\arraystretch}{1.3}
    \setlength{\tabcolsep}{2pt}
    \begin{tabular}{|r|ccccccc|c|}
    \cline{2-9}
    \multicolumn{1}{c|}{} & \textbf{DS}\textsubscript{en} & \textbf{MPQA}\textsubscript{en} & \textbf{MB}\textsubscript{ca} & \textbf{MB}\textsubscript{eu} & \textbf{NoReC}\textsubscript{no} & \textbf{OpeNER}\textsubscript{en} & \textbf{OpeNER}\textsubscript{es } & $\mu$\\  
    \hline 
    *\cite{lin-etal-2022-zhixiaobao}    & \textbf{49.4} & \textbf{44.7} & 72.8 & \textbf{73.9} & 52.9 & 76.0 & 72.2 & 63.1 \\
    *\cite{chen-etal-2022-mt}           & 48.5 & 41.6 & 72.8 & \textbf{73.9} & 52.4 & \textbf{76.3} & 74.2 & 62.8 \\
    *\cite{morio-etal-2022-hitachi}     & 46.3 & 40.2 & 70.9 & 71.5 & \textbf{53.3} & 75.6 & 73.2 & 61.6 \\
    *\cite{barikbin-2022-slpl}          & 41.0 & 37.5 & 68.1 & 72.3 & 50.4 & 74.7 & 73.5 & 59.6 \\
    *\cite{alonso-alonso-etal-2022-lys}  & 41.4 & 34.9 & 65.3 & 68.0 & 46.2 & 69.8 & 69.2 & 56.4 \\
    *\cite{zhang-etal-2022-ecnu}         & 49.0 & 35.1 & 68.4 & 68.6 & 49.6 & 67.6 & 62.3 & 56.1 \\
    \cite{barnes-etal-2021-structured} & 20.4 & 12.5 & 51.6 & 54.5 & 27.2 & 52.1 & 49.5 & 38.3 \\
     \cite{barnes-etal-2022-semeval}    & 0.06 & 0.02 & 33.8 & 36.5 & 12.3 & 32.9 & 24.0 & 19.9 \\
     SFA & 27.4 & 20.4 & 59.3 & 62.7 & 32.7 & - & - & 44.3\\
     PERIN     & 31.2 & 34.1 & 63.3 & 61.3 & 41.6 & - & - & 46.3 \\
     \cite{li2025extracting} & 34.1 & 24.1 & 63.2 & 62.6 & 41.9 & - & - & 45.9 \\
     \cite{fernandez2026structured} & 29.9 & 22.2 & 61.0 & 62.0 & 35.8 & - & - & 42.9 \\
    \hline 
    \textbf{Ours}  & 46.1 & 28.8 & \textbf{77.0} & 68.6 & 51.9 & 75.7 & \textbf{74.8} & 60.4 \\
    \hline
    \end{tabular}
    }
\end{table}

\paragraph{Discussion on head-first results}

Regarding the encodings, under the \textit{h.first} setting, the absolute encoding outperforms the biaffine baseline on 6 out of 7 treebanks (all except MPQA\textsubscript{en}), and its overall average is also higher (55.6 SF\textsubscript{1} vs. 51.8 for biaffine).The relative encoding, however, does not outperform the biaffine baseline on any treebank and has a lower overall average (44.1 SF\textsubscript{1}).

The bracketing encoding (k=3) achieves an average SF\textsubscript{1} of 55.5, while the bounded encodings consistently surpass the biaffine baseline across treebanks. In particular, the 6k-bit encoding (k=2) reaches an average across treebanks of 60.4, the highest overall,
suggesting that the choice of representation plays an important role. The choice of the hyperparameter k also has an impact on performance. When increasing k leads to higher graph coverage, performance generally improves. For example, in MB\textsubscript{eu}, increasing k from 1 to 3 in the bracketing encoding raises SF\textsubscript{1} from 60.5 to 69.5. By contrast, once coverage gains saturate, further increases in k provide little or no benefit. This is illustrated by DS\textsubscript{en}, where increasing k from 2 to 3 for the 6k-bit encoding reduces SF\textsubscript{1} from 46.1 to 43.3, consistent with the limited coverage improvement reported in Table~\ref{tab:coverage}.

\begin{figure}
	\centering
		\includegraphics[width=\linewidth]{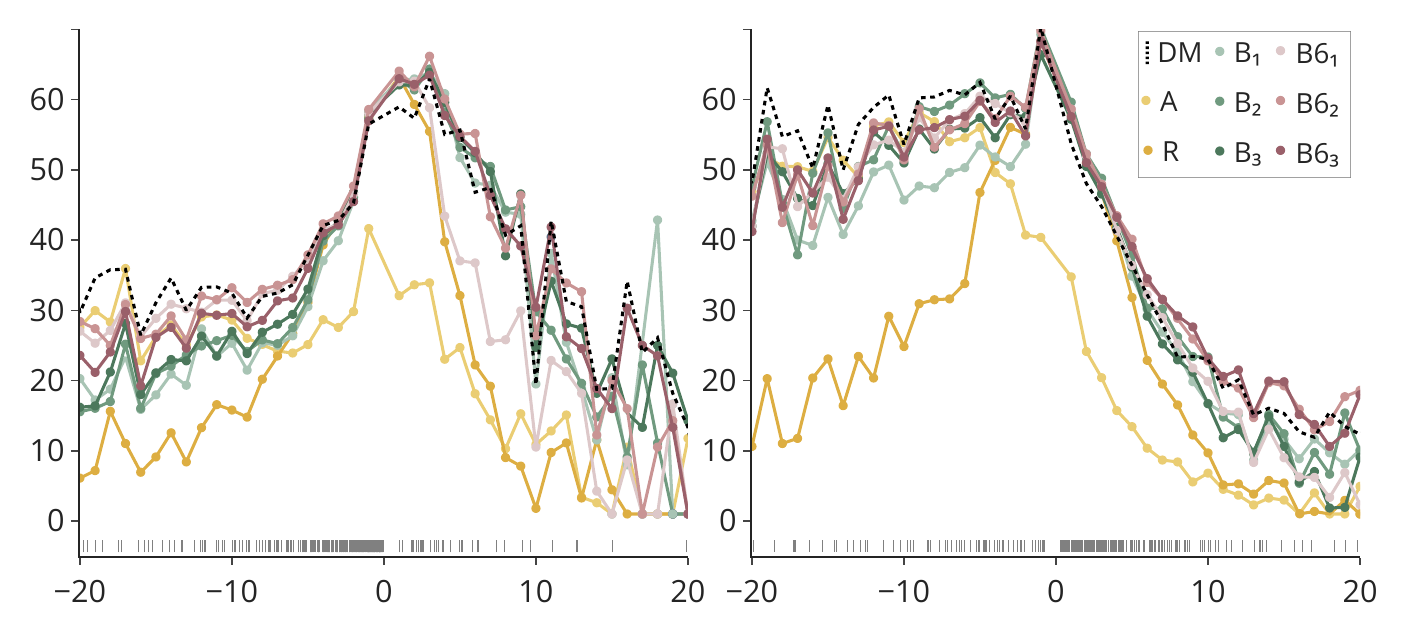}
	\caption{F\textsubscript{1}-score ($y$-axis) per displacement ($x$-axis) in the head-first (left) and head-final (right) representations, alongside the rug plot of the ground truth displacements. The biaffine baseline (DM) and the encodings are denoted with different acronyms: absolute (A), relative (R), bracketing (B), and $6k$-bit (B6) encodings, using subscripts for the value of the hyperparameter $k$, when applicable.}
	\label{fig:displacements}
\end{figure}

\paragraph{Discussion on head-final results}

Comparing the \textit{h.first} and \textit{h.final} transformations reveals distinct modeling behaviors. While SL models achieve stronger results under the \textit{h.first} setting, the \textit{h.final} representation appears more challenging for sequence labeling models, leading to lower performance across encodings. In contrast, the biaffine baseline achieves its best results under the \textit{h.final} setting, suggesting that this representation is better suited to graph-based decoding. Nevertheless, the $6k$-bit SL model with \textit{h.first} remains the best-performing approach overall, obtaining average SF\textsubscript{1} scores of 59.4, 60.4, and 59.2 for $k = 1$, 2, and 3, respectively, and surpassing the best biaffine result of 57.8.

Complementarily, Figure \ref{fig:displacements} presents a detailed visualization of the F\textsubscript{1}-score across different displacement values for the \textit{h.first} and \textit{h.final} representations. The displacement of a syntactic dependency is a signed distance, i.e., the position of the head word in the sentence minus that of the dependent word. For both graph-based and sequence-labeling parsers, all models show lower performance for longer dependencies, regardless of their direction, reflecting a general difficulty in capturing long-range dependencies across paradigms. Although the displacement distribution in the \textit{h.first} setting is skewed to the right, indicating a higher frequency of negative displacements (right arcs), the various encodings, particularly the $6k$-bit encodings, maintain relatively balanced performance across positive and negative displacements. This suggests that \textit{h.first} is more robust to the skew in the data and provides more consistent performance across dependency lengths and directions. In contrast, the \textit{h.final} setting exhibits a left-skewed distribution, meaning that the training data contains more positive displacements (left arcs). Its performance is notably lower on the denser positive-displacement side, which negatively affects the overall accuracy of the model.

Finally, Table~\ref{tab:results-comparison} presents a comparison between our approach and representative existing systems for structured sentiment analysis. While previous approaches often incorporate additional components, such as data augmentation \cite{chen-etal-2022-mt}, ensemble strategies \cite{barikbin-2022-slpl}, or language-specific encoders \cite{lin-etal-2022-zhixiaobao,morio-etal-2022-hitachi}, our approach relies on a single encoder architecture shared across datasets, without introducing treebank-specific optimizations. Moreover, it formulates structured sentiment analysis as a sequence labeling problem, avoiding graph-specific decoding mechanisms. Despite this relatively simple formulation, our parser outperforms existing systems on 2 out of 7 treebanks and achieves competitive performance across the seven treebanks. These results suggest that formulating graph parsing as sequence labeling can provide a competitive yet simple alternative for structured sentiment analysis.

\subsection{Fine-grained Evaluation} \label{fine-grained-evaluation}

\begin{table}[h]\centering\footnotesize
         \caption{\imran{Fine-grained SSA results comparison against state of the art systems and Biaffine baseline: \textbf{Spans (F\textsubscript{1})} column is the token-level F\textsubscript{1}-score for the holder, target and expression; \textbf{T.SF\textsubscript{1}} is the targeted F\textsubscript{1} score; \textbf{UF\textsubscript{1}} and\textbf{ LF\textsubscript{1}} are the unlabeled and labeled F\textsubscript{1} scores for dependency graph parsing; and S\textbf{F\textsubscript{1}} and \textbf{NSF\textsubscript{1}} are the sentiment F\textsubscript{1} score with or without the sentiment polarity. Our best-performing encoding $6k$-bit (\textbf{B6}\textsubscript{3}) where subscript denotes the value of the hyperparemeter $k$. Dash (-) for results not reported in the original paper. Best result highlighted in bold.}}\label{tab:fine-grained-results} 
    \resizebox{\columnwidth}{!}{%
    \setlength{\tabcolsep}{4pt}
    \renewcommand{\arraystretch}{1.1}
    \begin{tabular}{|l|l|c|ccc|c|cc|cc|}
        \hline
         \multicolumn{1}{|c|}{\textbf{Dataset}} & \multicolumn{2}{c|}{\textbf{Model}} & \multicolumn{3}{c|}{\textbf{Spans (F\textsubscript{1})}} & \multirow{2}{*}{\textbf{T.SF\textsubscript{1}}} & \multicolumn{2}{c|}{\textbf{Dep.}} & \multicolumn{2}{c|}{\textbf{Sent.}} \\ 
         \cline{4-6}\cline{8-11}
         \multicolumn{1}{|c|}{} & \multicolumn{2}{c|}{} & Holder & Target & Exp & & UF\textsubscript{1} & LF\textsubscript{1} & NSF\textsubscript{1} & SF\textsubscript{1} \\ 
         \hline
         \multirow{7}{*}{\textbf{MB}\textsubscript{ca}} & \multicolumn{2}{|c|}{\cite{samuel-etal-2022-direct}\textsuperscript{OT}} & 48.0 & 72.5 & 68.9 & - & - & - & 65.7 & 63.3 \\

        & \multicolumn{2}{|c|}{\cite{samuel-etal-2022-direct}\textsuperscript{LE}} &\textbf{60.8} & 70.8 & 72.5 & - & - & - & 64.5 & 62.1 \\
        & \multicolumn{2}{|c|}{\cite{samuel-etal-2022-direct}\textsuperscript{NC}} & 56.1 & 69.8 & 70.5 & - & - & - & 63.5 & 61.7 \\
        & \multicolumn{2}{|c|}{\cite{li2025extracting}} & 36.8 & - & \textbf{75.5} & \textbf{73.4} & - & - & 66.4 & 63.2 \\
        & \multicolumn{2}{|c|}{\cite{peng2021sparse}} &  46.2 & 74.2 & 71.0 &  60.9 & & 64.5 & - & 59.3 \\
        
        & \multicolumn{2}{|c|}{\cite{barnes-etal-2021-structured}} & 37.1 & 71.2 & 67.1 & 53.9 &  62.7 & 58.1 & 59.7 & 53.7\\ 
        & \multicolumn{2}{|c|}{\cite{fernandez2026structured}} & 55.7 & 73.9 & 69.6 & 57.0 & 65.2 & 60.2 & 64.7 & 59.0 \\
         
         & \multicolumn{2}{|c|}{Biaf} & 54.5 & 74.2 & 71.1 & 50.8 & 50.8 & 49.4 & 46.6 & 45.1 \\
        \cline{2-11}
        & \multicolumn{2}{|c|}{\textbf{B6}\textsubscript{3}} & 56.3 & \textbf{80.2} & 74.5 & 67.2 & \textbf{69.3} & \textbf{67.0} & \textbf{67.6} & \textbf{66.3} \\
        \hline
        \hline
        \multirow{7}{*}{\textbf{MB}\textsubscript{eu}} & \multicolumn{2}{|c|}{\cite{peng2021sparse}} & 65.8 & 71.0 & 76.7 & 59.6 & - & \textbf{66.1} & - & 62.7 \\

        & \multicolumn{2}{|c|}{\cite{samuel-etal-2022-direct}\textsuperscript{NC}} & 58.9 & 63.5 & 73.9 & - & - & - & 59.8 & 58.6 \\
        & \multicolumn{2}{|c|}{\cite{samuel-etal-2022-direct}\textsuperscript{LE}} &57.6 & 64.9 & 72.5 & - & - & - & 60.0 & 58.8 \\
        & \multicolumn{2}{|c|}{\cite{samuel-etal-2022-direct}\textsuperscript{OT}} &  64.2 & 67.4 & 73.2 & - & - & - & 62.5 & 61.3 \\
        & \multicolumn{2}{|c|}{\cite{li2025extracting}} & 61.4 & - & 70.0 & \textbf{74.5} & - & - & 63.8 & 62.6 \\

        & \multicolumn{2}{|c|}{\cite{barnes-etal-2021-structured}} & 60.5 & 64.0 & 72.1 & 56.9 & 60.8 & 56.0 & 58.0 & 54.7 \\
        & \multicolumn{2}{|c|}{\cite{fernandez2026structured}} & 66.6 & 67.9 & 76.6 & 60.5 & \textbf{66.9} & 61.4 & 63.8 & 60.4 \\
        
        & \multicolumn{2}{|c|}{Biaf} & 71.4 & 74.4 & 78.2 & 55.0 & 57.0 & 55.6 & 44.4 & 44.0 \\
        \cline{2-11}
        & \multicolumn{2}{|c|}{\textbf{B6}\textsubscript{3}} & \textbf{71.8} & \textbf{76.1} & \textbf{78.9} & 66.8 & 66.5 & 64.3 & \textbf{65.1} & \textbf{63.8} \\
        \hline
        \hline
        \multirow{7}{*}{\textbf{NoReC}\textsubscript{no}} & \multicolumn{2}{|c|}{\cite{li2025extracting}} & 62.5 & - & 61.4 & \textbf{61.3} & - & - & 46.7 & 41.9 \\

        & \multicolumn{2}{|c|}{\cite{samuel-etal-2022-direct}\textsuperscript{NC}} & 60.3 & 51.8 & 54.2 & - & - & - & 42.7 & 39.3 \\
        & \multicolumn{2}{|c|}{\cite{samuel-etal-2022-direct}\textsuperscript{LE}} & 64.0 & 52.3 & 56.1 & - & - & - & 43.7 & 40.4 \\
        & \multicolumn{2}{|c|}{\cite{samuel-etal-2022-direct}\textsuperscript{OT}} & \textbf{65.1} & 58.3 & 60.7 & - & - & - & \textbf{47.8} & 41.6 \\
        & \multicolumn{2}{|c|}{\cite{peng2021sparse}} & 63.6 & 55.3 & 56.1 &  31.5 & &  40.0 & - & 32.7 \\

        & \multicolumn{2}{|c|}{\cite{barnes-etal-2021-structured}} & 60.4 & 54.8 & 55.5 & 31.9 & 48.0 & 37.7 & 39.2 & 31.2 \\
        & \multicolumn{2}{|c|}{\cite{fernandez2026structured}} & 57.5 & 54.3 & 51.5 & 36.1 & 45.0 & 36.5 & 44.2 & 35.8 \\

        &  \multicolumn{2}{|c|}{Biaf} & 64.2 & 61.8 & 61.3 & 33.8 & 30.3 & 27.3 & 28.2 & 26.6 \\
        \cline{2-11}
        &  \multicolumn{2}{|c|}{\textbf{B6}\textsubscript{3}} & 62.8 & \textbf{62.0} & \textbf{66.1} & 47.5 & \textbf{55.9} & \textbf{50.5} & 47.4 & \textbf{43.8} \\
        \hline
        \hline
        \multirow{7}{*}{\textbf{DS}\textsubscript{en}} & \multicolumn{2}{|c|}{\cite{li2025extracting}} & 47.0 & - & 44.7 & \textbf{45.0} & - & - & \textbf{37.8} & \textbf{34.1} \\

        & \multicolumn{2}{|c|}{\cite{peng2021sparse}} & \textbf{50.0} & 44.8 & 43.7 & 30.7 & - & 34.9 & - & 27.4 \\
        & \multicolumn{2}{|c|}{\cite{samuel-etal-2022-direct}\textsuperscript{NC}} &  31.4 & 35.0 & 35.1 & - & - & - & 24.8 & 22.9 \\
        & \multicolumn{2}{|c|}{\cite{samuel-etal-2022-direct}\textsuperscript{LE}} &32.5 & 38.0 & 36.2 & - & - & - & 28.8 & 27.3  \\
        & \multicolumn{2}{|c|}{\cite{samuel-etal-2022-direct}\textsuperscript{OT}} &  42.2 & 40.6 & 39.3 & - & - & - & 33.2 & 31.2\\

        & \multicolumn{2}{|c|}{\cite{barnes-etal-2021-structured}} &  37.4 & 42.1 & 45.5 & 29.6 & 38.1 & 33.9 & 34.3 & 26.5\\
        & \multicolumn{2}{|c|}{\cite{fernandez2026structured}} & 43.2 & 44.4 & 43.6 & 28.7 & 36.3 & 30.3 & 32.7 & 25.6 \\
        
        & \multicolumn{2}{|c|}{Biaf} & 37.5 & 35.3 & 43.8 & 18.0 & 23.4 & 21.7 & 9.7 & 9.7 \\
        \cline{2-11}
        & \multicolumn{2}{|c|}{\textbf{B6}\textsubscript{3}} & 28.6 & \textbf{45.6} & \textbf{53.0} & 34.3 & \textbf{45.8} & \textbf{42.7} & 36.1 & 32.0 \\
         \cline{3-11}
        \hline
        \hline
        \multirow{7}{*}{\textbf{MPQA}\textsubscript{en}} & \multicolumn{2}{|c|}{\cite{samuel-etal-2022-direct}\textsuperscript{OT}} & 55.7 & \textbf{64.0} & 53.5 & - & - & - & \textbf{45.1} & \textbf{34.1} \\
        & \multicolumn{2}{|c|}{\cite{samuel-etal-2022-direct}\textsuperscript{NC}} & \textbf{58.4} & 60.3 & \textbf{55.8} & - & - & - & 38.7 & 28.3  \\

        & \multicolumn{2}{|c|}{\cite{samuel-etal-2022-direct}\textsuperscript{LE}} &  53.6 & 53.4 & 53.4 & - & - & - & 33.8 & 27.0 \\
        & \multicolumn{2}{|c|}{\cite{peng2021sparse}} & 47.9 & 50.7 & 47.8 & 33.7 & & 37.6 & - & 20.4  \\
        & \multicolumn{2}{|c|}{\cite{li2025extracting}} & 46.5 & - & 51.8 & \textbf{47.6} & - & - & 29.3 & 24.1 \\

        & \multicolumn{2}{|c|}{\cite{barnes-etal-2021-structured}} & 46.3 &   49.5 &  46.0  & 18.6 &  41.4 &  38.0  & 26.1 &  18.8 \\
        & \multicolumn{2}{|c|}{\cite{fernandez2026structured}} & 43.8 & 46.4 & 45.2 & 18.1 & 38.0 & 35.4 & 25.9 & 19.7 \\

        
        & \multicolumn{2}{|c|}{Biaf} & 52.4 & 55.5 & 53.0 & 29.0 & 25.2 & 24.0 & 14.0 & 12.0 \\
        \cline{2-11}
        & \multicolumn{2}{|c|}{\textbf{B6}\textsubscript{3}} & 55.6 & 61.0 & 54.5 & 26.0 & \textbf{50.2} & \textbf{47.8} & 29.9 & 27.3 \\
        \cline{3-11}
        \hline
    \end{tabular}%
    }
\end{table}

Following Protocol~2, Table~\ref{tab:fine-grained-results} compares our sequence labeling approach with prior work evaluated under the same strategy (per-dataset results are in Tables~\ref{tab:mb-ca-results}--\ref{tab:opener-es-results}, Appendix~\ref{app:efficiency-tables}). These systems rely on alternative paradigms:
SFA \cite{peng2021sparse}, which models token-level sentiment relations with sparse fuzzy attention;
PERIN \cite{samuel-etal-2022-direct}, a graph parser that directly learns sentiment graph edges under three encodings\footnote{Node-centric (NC), labeled-edge (LE), and opinion-tuple (OT).};
\citet{barnes-etal-2021-structured}, who cast the task as dependency graph parsing with a dedicated graph decoder;
\citet{fernandez2026structured}, who use transition-based graph parsing;
and \citet{li2025extracting}, who extract (holder, target, opinion, polarity) quadruples with a boundary-sensitive token-pair labeling framework.

Three main conclusions emerge from Table~\ref{tab:fine-grained-results}, following the results from the previous section. First, they confirm that sentiment analysis does not require a dedicated graph-parsing architecture. With a suitable linearization, a plain sequence labeler obtains the highest SF\textsubscript{1} among the available systems on three of the five benchmarks (MB\textsubscript{ca}, MB\textsubscript{eu} and NoReC\textsubscript{no}), improving over +1.1 to +3.0 SF$_1$ points (largest gain on MB\textsubscript{ca}), as well as the best target F\textsubscript{1} on four of them. On the remaining two datasets it stays competitive with systems built specifically to predict graphs or opinion tuples.

Second, the output representation, not only the encoder, drives these gains. Particularly, the biaffine baseline uses the same XLM-RoBERTa-large encoder and extracts spans about as well as our model, sometimes better, yet its sentiment graph scores fall far behind on every dataset---by over 15 SF$_1$ points. The encoder therefore already identifies the relevant text fragments; the difficulty lies in assembling them into coherent graphs, which is precisely the step our linearization shows can be performed without a dedicated decoder.

Thus, one of our model's strengths lies in capturing complete opinion structures rather than individual elements. It is most competitive on graph-level metrics, while element-level metrics, especially holder extraction and targeted polarity, are where other systems often do better. We hypothesize that this is consistent with an encoding in which each token's label encodes its role within the full opinion graph: the model learns global structure well, at some cost in precision on individual components. Improving holder and target-polarity prediction is therefore the most direct path to further gains, although it falls outside the scope of a pure sequence-labeling-based approach.

Finally, the gap to the best available systems on MPQA\textsubscript{en} and DS\textsubscript{en} reflects the difficulty of these datasets more than a specific weakness of our formulation. Following Table \ref{tab:dataset_stats}, they have sparse, long-range and frequently overlapping opinions, and all systems score substantially lower on them. A suitable explanation for this is that set-based decoders such as PERIN's opinion-tuple variant, which predict each opinion as an independent unit, seem better suited to heavily overlapping structures.

\subsection{Model Efficiency Analysis}

Our sequence labeling approach is substantially more efficient than the Biaffine graph decoder. Using the \textit{head-first} representation (\textit{head-final} exhibits similar patterns), our models process approximately 3,800 tokens/s during training and 4,450 tokens/s at inference—a 2$\times$ speedup over the Biaffine baseline (1,893 and 2,196 tokens/s, respectively) under the same configurations (e.g., batch size=64)\footnote{All throughput measurements (training and inference) were performed on a single NVIDIA RTX 6000 Ada GPU.}. These efficiency gains come directly from the simpler sequence-labeling decoder, which avoids constructing dense pairwise score tensors. As shown in Table~\ref{tab:model-efficiency-analysis} of Appendix~\ref{app:efficiency-tables}, all representation variants (A, R, {B}\textsubscript{1}--{B}\textsubscript{3}, and {B6}\textsubscript{1}--{B6}\textsubscript{3}) perform within 1--2\% of each other, confirming that the advantage stems from the formulation itself rather than any specific encoding.

\begin{figure}
	\centering
		\includegraphics[width=\linewidth]{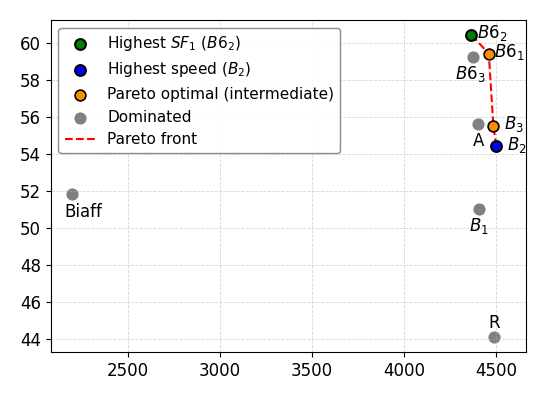}
	\caption{Performance (SF\textsubscript{1}, y-axis) vs inference speed
(tokens per second, x-axis) on the head-first representation. The Pareto front is displayed with dashed lines. The biaffine baseline (Biaff) and the encodings are denoted with different acronyms: absolute (A), relative (R), bracketing (B), and $6k$-bit (B6), using subscripts for the value of the hyperparameter $k$, where applicable.}
	\label{fig:pareto-front}
\end{figure}

Figure~\ref{fig:pareto-front} shows an efficiency comparison of our models in terms of inference speed (tokens per second, Table \ref{tab:model-efficiency-analysis}) vs.\ $SF_1$ (Table \ref{tab:results}) on the \textit{head-first} representation, highlighting the Pareto front with dashed lines. As can be seen, all sequence labeling models are substantially faster than the biaffine baseline parser. The bracketing encodings ($B_2$ and $B_3$) occupy the high-speed end of the front, whereas the $6k$-bit encodings ($B6_1$ and $B6_2$) achieve the higher $SF_1$ scores---likely due to the compact label space (Table \ref{tab:label-space}) making classification easier at the cost of a slight drop in speed.

\section{Conclusion} \label{Conclusion}

This work framed structured sentiment analysis as a dependency graph parsing problem. Unlike previous attempts, it applied the parsing-as-tagging paradigm, where each input token is assigned a  label that encodes a part of the graph. To achieve this, we leveraged a recently proposed approach to linearize graphs, applying it to graph-based sentiment structures.
We showed that these linearizations are well-suited for the task at hand and that both bounded and unbounded encodings (in terms of their output label space) are learnable.

Particularly, when these are combined with standard encoder-decoder sequence architectures and the right representation, sequence labeling models can consistently compete with or even outperform more complex approaches across different datasets, while achieving high efficiency.
Of the linearizations tested, the 6k-bit encoding showed the best overall performance. Overall, these findings highlight the potential of lightweight, sequence-labeling models for structured sentiment analysis, especially when they are combined with effective structural encodings.

\section*{Acknowledgments}
We acknowledge the European Research Council (ERC), which has funded this research under the Horizon Europe research and innovation programme (SALSA, grant agreement No 101100615), 
GAP (PID2022-139308OA-I00) funded by MICIU/AEI/10.13039/501100011033/ and by ERDF, EU; LATCHING (PID2023-147129OB-C21) funded by MICIU/AEI/10.13039/501100011033 and ERDF (EU), Ministry for Digital Transformation and Civil Service and “NextGenerationEU” PRTR under grant TSI-100925-2023-1, Xunta de Galicia (ED431C 2024/02),  and Centro de Investigación de Galicia ‘‘CITIC’’, funded by the Xunta de Galicia through the collaboration agreement between the Consellería de Cultura, Educación, Formación Profesional e Universidades and the Galician universities for the reinforcement of the research centres of the Galician University System (CIGUS). Furthermore, this research was supported by the International, Interdisciplinary and Intersectoral Information and Communications Technology PhD programme (3-i ICT) granted to CITIC and supported by the European Union through the Horizon 2020 research and innovation programme under a Marie Skłodowska-Curie agreement (H2020-MSCA-COFUND), GA 101034261. Additional support was provided by the INDITEX-UDC 2024 pre-doctoral Research Fellowship Program.


\printcredits

\bibliographystyle{model1-num-names}

\bibliography{references}

\newpage
\bio{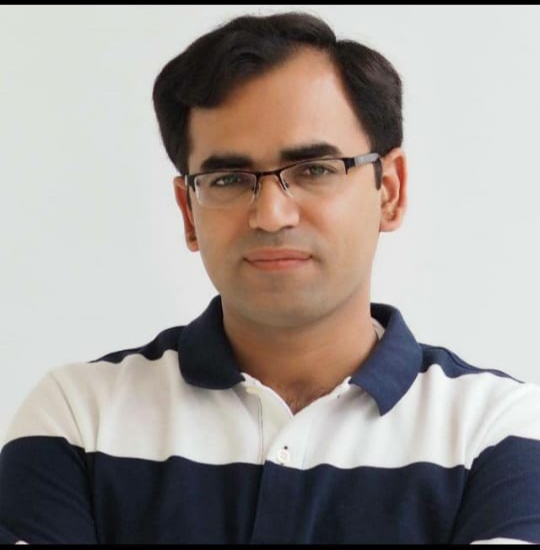}
Muhammad Imran is a Marie Sklodowska-Curie Doctoral Fellow in the 3-i ICT Programme at the University of A Coruña, Spain. He joined the Natural Language Processing Group (LyS-Group) at the university of A Coruña in 2023. His research work is focused on Sequence Labelling Parsing for Applied Natural Language Processing. His research interests include Machine Learning and its applications to Natural Language Processing (NLP), particularly Sentiment Analysis, Named Entity Recognition, syntactic parsing (Dependency and Constituency), and graph-based parsing methods. He has an industrial background given his experience (9+ Years) as a Senior Software Engineer in USA-based IT organizations where he contributed to the development of various cutting-edge software applications.
\endbio

\bio{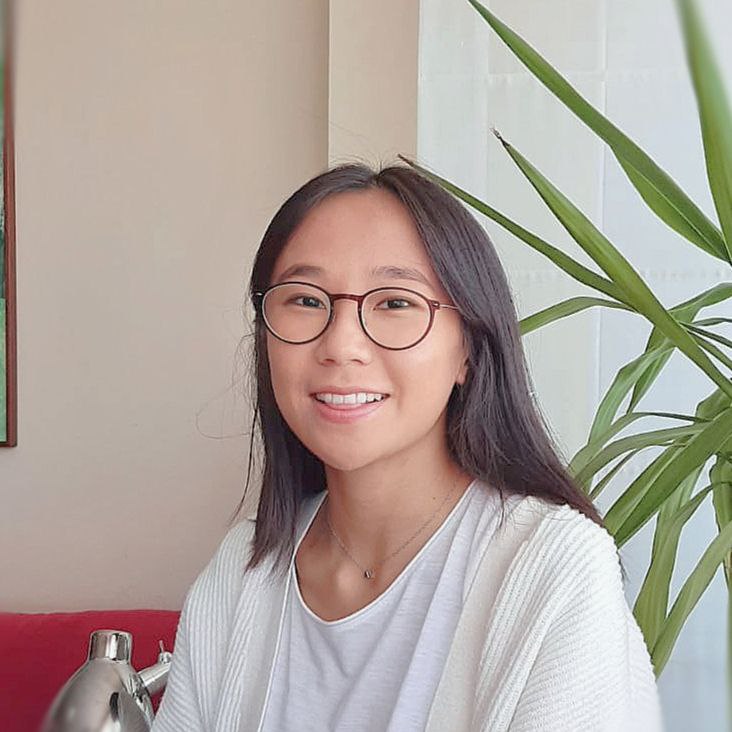}
Ana Ezquerro is a PhD Candidate at the Graz University of Technology (TU Graz). She holds a B.Sc. in Data Science and Engineering (2019–2023) and a recently completed M.Sc. in Artificial Intelligence (2023–2025) from University of A Coruña (UDC). Her research has focused on Natural Language Processing (NLP), with particular emphasis on dependency, constituency, and graph parsing; and multimodal learning.
\endbio

\bio{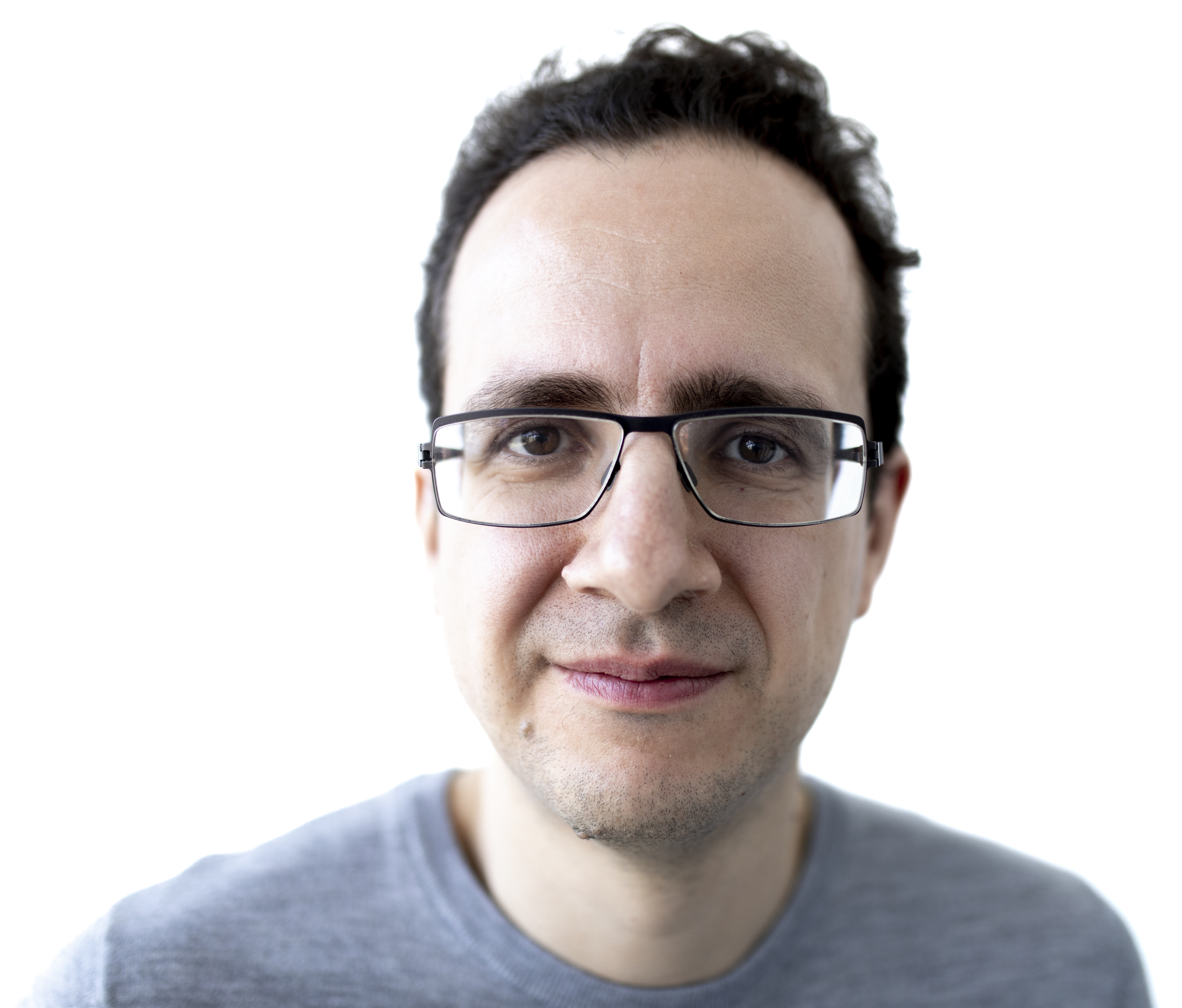}
Carlos Gómez-Rodríguez is an Engineer (2005) and PhD in Computer Science (2009) from the Universidade da Coruña (UDC), and is currently a Full Professor in the area of Computer Science and Artificial Intelligence at said institution.

His research focuses on the field of computational linguistics, especially syntactic analysis and opinion mining, having authored more than 150 peer-reviewed publications, including 40 in top-level conferences (Class 1 in the GGS classification) and 39 in JCR-indexed journals. He was the principal investigator of an ERC Starting Grant project and now leads an ERC Proof-of-Concept Grant (SALSA: Syntactic Analysis for Large-Scale Sentiment Analysis), apart from various national and regional-level research projects. 

He has supervised 5 completed doctoral theses with another two in progress. In 2022, he received the "María Andresa Casamayor" National Youth Research Award from the Spanish Ministry of Science, the highest recognition at the Spanish level for young researchers in Mathematics and Information and Communication Technologies. He was also recognized in the same year as Honorary Member of the Royal Spanish Mathematical Society.
\endbio

\bio{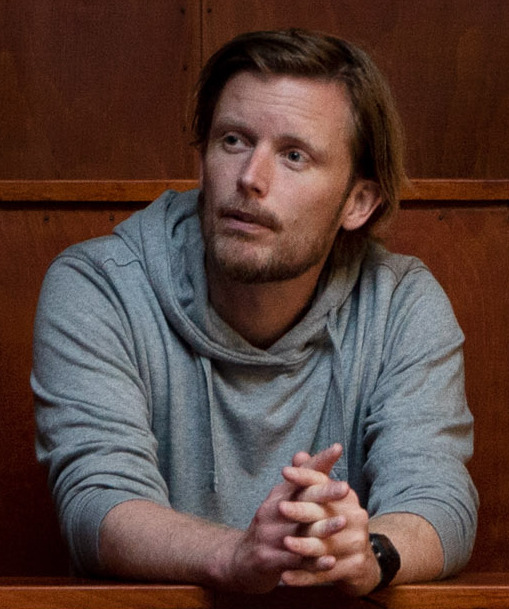}
Anders Søgaard is a Full Professor of NLP and Machine Learning at the University of Copenhagen. He previously worked for University of Potsdam and Google Research. He was awarded many best paper awards at top conferences and several prestigious grants, including ERC Starting Grant, Google Focused Research Award, and Carlsberg Semper Ardens Advanced. He is a Chief Scientist at the Center for AI in Society and leads the Center for Philosophy of AI, both at the University of Copenhagen.
\endbio

\newpage
\bio{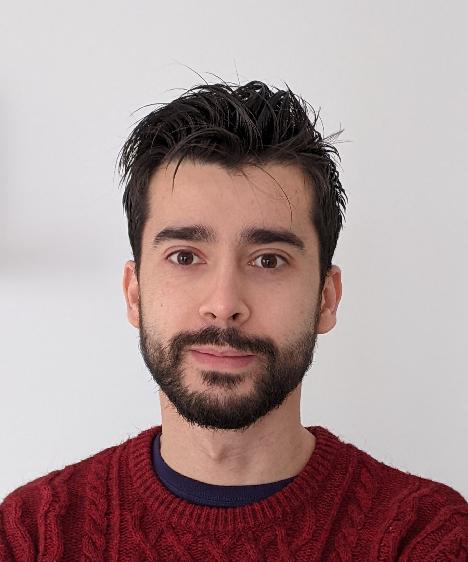}
David Vilares is an Assistant Professor in Computer Science and Artificial Intelligence at Universidade da Coruña (UDC). His research lies at the intersection of machine learning and computational linguistics, covering topics such as structured prediction, low-resource languages, and computational social science. In particular, he has contributed both theoretically and practically to compositional methods for understanding language, sequence labeling approaches for the efficient processing of tree- and graph-structured data, and the use of NLP tools to analyze social behavior on social media. His work has been published in 12 JCR-indexed journals and 16 top-tier conferences (Class 1 in the GGS classification), including ACL, EMNLP, and NAACL. He was the principal investigator of two national projects. He has also supervised one doctoral thesis, with two more currently in progress. In 2018, he received the National Research Award for Young Researchers in Computer Science, granted by the Spanish Society of Computer Science and the BBVA Foundation.
\endbio

\clearpage 
\setcounter{table}{0}
\renewcommand{\thetable}{A\arabic{table}}   

\appendix
\section{Appendix} \label{app:efficiency-tables}
The detailed fine-grained SSA results for each dataset are provided in Tables~\ref{tab:mb-ca-results} through~\ref{tab:opener-es-results}. Table~\ref{tab:model-efficiency-analysis} provides model efficiency in terms of training throughput and inference throughput (tokens/s and sentences/s). All encodings: Absolute (A), Relative (R), Bracketing ({B}\textsubscript{1}--{B}\textsubscript{3}), and  $6k$-bit ({B6}\textsubscript{1}--{B6}\textsubscript{3}) are compared against the Biaffine (Biaf) graph-based baseline.

\captionsetup[table]{
  justification = raggedright,   
  singlelinecheck = false,       
  labelfont = bf,                
  labelsep = newline,            
  skip = 12pt,
}

\makeatletter
\long\def\@makecaption#1#2{%
  \vskip\abovecaptionskip
  \sbox\@tempboxa{\bfseries #1}
  \noindent\usebox\@tempboxa\par       
  \noindent #2\par                     
  \vskip\belowcaptionskip
}
\makeatother

 \captionsetup{type=table} 
  \caption{}{Fine-grained results on the MultiBooked (Catalan) dataset. \textbf{Spans (F\textsubscript{1})} column is the token-level F\textsubscript{1}-score for the holder, target and expression; \textbf{T.SF\textsubscript{1}} is the targeted F\textsubscript{1} score; \textbf{UF\textsubscript{1}} and\textbf{ LF\textsubscript{1}} are the unlabeled and labeled F\textsubscript{1} scores for dependency graph parsing; and S\textbf{F\textsubscript{1}} and \textbf{NSF\textsubscript{1}} are the sentiment F\textsubscript{1} score with or without the sentiment polarity. Each encoding is abbreviated with acronyms: absolute (A), relative (R), bracketing (B) and $6k$-bit, using subscripts to denote the value of the hyperparemeter $k$. Dash (-) for results not reported in the original paper. Best result highlighted in bold.} \label{tab:mb-ca-results}
  \vspace{5pt}
    \resizebox{\columnwidth}{!}{
    \setlength{\tabcolsep}{4pt}
    \renewcommand{\arraystretch}{1.1}
    \begin{tabular}{|l|c|ccc|c|cc|cc|}
        \cline{3-10} 
         \multicolumn{2}{c|}{}& \multicolumn{3}{c|}{\textbf{Spans (F\textsubscript{1})}} & \multirow{2}{*}{\textbf{T.SF\textsubscript{1}}} & \multicolumn{2}{c|}{\textbf{Dep.}} & \multicolumn{2}{c|}{\textbf{Sent.}} \\ 
         \cline{3-5}\cline{7-10}
         \multicolumn{2}{c|}{}& Holder & Target & Exp & & UF\textsubscript{1} & LF\textsubscript{1} & NSF\textsubscript{1} & SF\textsubscript{1} \\ 
         \hline 
        \multicolumn{2}{|c|}{\cite{peng2021sparse}} &  46.2 & 74.2 & 71.0 &  60.9 & & 64.5 & - & 59.3 \\
        \multicolumn{2}{|c|}{\cite{samuel-etal-2022-direct}\textsuperscript{NC}} & 56.1 & 69.8 & 70.5 & - & - & - & 63.5 & 61.7 \\
        \multicolumn{2}{|c|}{\cite{samuel-etal-2022-direct}\textsuperscript{LE}} &60.8 & 70.8 & 72.5 & - & - & - & 64.5 & 62.1 \\
        \multicolumn{2}{|c|}{\cite{samuel-etal-2022-direct}\textsuperscript{OT}} & 48.0 & 72.5 & 68.9 & & & & 65.7 & 63.3  \\
        \multicolumn{2}{|c|}{\cite{li2025extracting}} & 36.8 & - & 75.5 & 73.4 & - & - & 66.4 & 63.2 \\
        \hline
        \parbox[t]{2mm}{\multirow{13}{*}{\rotatebox[origin=c]{90}{\textit{h.first}}}} 
            & \textbf{A} & 64.0 & 75.8 & 74.5 & 51.7 & 50.8 & 49.2 & 47.0 & 45.5 \\ 
            & \textbf{R} & 60.3 & 77.7 & 73.5 & 60.4 & 62.2 & 59.8 & 52.0 & 50.4 \\
            & \textbf{B}\textsubscript{1} & 60.9 & 78.9 & 74.6 & 69.8 & 65.1 & 63.3 & 51.0 & 49.7 \\
            & \textbf{B}\textsubscript{2} & 55.6 & 79.1 & 76.8 & 69.8 & 66.9 & 65.1 & 60.4 & 59.0  \\
            & \textbf{B}\textsubscript{3} & 50.9 & 77.2 & 75.4 & 67.2 & 65.2 & 63.5 & 56.5 & 55.2 \\
            & \textbf{B6}\textsubscript{1} & 55.7 & 79.7 & 75.2 & \textbf{73.0} & 68.0 & 66.1 & 57.7 & 56.6 \\
            & \textbf{B6}\textsubscript{2} & 60.0 & 79.9 & \textbf{79.4} & 72.0 & \textbf{70.8} & 68.7 & 69.5 & \textbf{67.8}\\
            & \textbf{B6}\textsubscript{3} & 56.3 & 79.2 & 75.8 & 71.3 & 70.0 & \textbf{68.3} & 66.0 & 64.2 \\
            \cline{2-10}
            & \cite{barnes-etal-2021-structured} &  43.0 & 72.5 & 71.1 & 55.0 & 66.8 & 62.1 & 62.0 & 56.8 \\
            & \cite{fernandez2026structured} & 54.1 & 75.4 & 71.4 & 64.0 & 68.5 & 64.4 & 67.0 & 61.0 \\
            & Biaf & 55.1 & 76.6 & 73.9 & 39.6 & 44.3 & 42.7 & 39.1 & 37.4 \\ 
        \hline 
        \parbox[t]{2mm}{\multirow{13}{*}{\rotatebox[origin=c]{90}{\textit{h.final}}}}  
            & \textbf{A} & 67.5 & 74.5 & 72.8 & 36.0  & 44.6 & 43.3 & 39.8 & 38.6 \\
            & \textbf{R} & 65.1 & 78.9 & 72.5 & 59.2 & 61.0 & 58.2 & 45.1 & 43.9 \\
            & \textbf{B}\textsubscript{1} & 54.6 & 77.1 & 70.5 & 58.4 & 67.1 & 64.6 & 58.3 & 56.0\\
            & \textbf{B}\textsubscript{2} &  58.0 & 75.4 & 72.3 & 62.0 & 67.7 & 65.7 & 66.8 & 66.0\\ 
            & \textbf{B}\textsubscript{3} & \textbf{67.6} & 75.4 & 73.3 & 58.5 & 69.0 & 65.4 & 62.8 & 61.4 \\
            & \textbf{B6}\textsubscript{1} & 54.5 & 79.1 & 75.4 & 64.1 & 68.5 & 65.7 & 59.7 & 58.2\\
            & \textbf{B6}\textsubscript{2} & 61.1 & 79.6 & 76.3 & 63.6 & 70.4 & 67.7 & \textbf{69.8} & 67.7 \\
            & \textbf{B6}\textsubscript{3} & 56.3 & \textbf{80.2} & 74.5 & 67.2 & 69.3 & 67.0 & 67.6 & 66.3\\
            \cline{2-10}
            & \cite{barnes-etal-2021-structured} & 37.1 & 71.2 & 67.1 & 53.9 &  62.7 & 58.1 & 59.7 & 53.7\\ 
            & \cite{fernandez2026structured} & 55.7 & 73.9 & 69.6 & 57.0 & 65.2 & 60.2 & 64.7 & 59.0 \\
            & Biaf & 54.5 & 74.2 & 71.1 & 50.8 & 50.8 & 49.4 & 46.6 & 45.1 \\ 
        \hline 
    \end{tabular}
    }
\begin{table}[h]\centering\footnotesize
    \caption{Fine-grained results on the MultiBooked (Basque) dataset. Same notation as in Table \ref{tab:mb-ca-results}.}\label{tab:mb-eu-results}
    \resizebox{\columnwidth}{!}{%
    \setlength{\tabcolsep}{4pt}
    \renewcommand{\arraystretch}{1.1}
    \begin{tabular}{|l|c|ccc|c|cc|cc|}
        \cline{3-10} 
         \multicolumn{2}{c|}{}& \multicolumn{3}{c|}{\textbf{Spans (F\textsubscript{1})}} & \multirow{2}{*}{\textbf{T.SF\textsubscript{1}}} & \multicolumn{2}{c|}{\textbf{Dep.}} & \multicolumn{2}{c|}{\textbf{Sent.}} \\ 
         \cline{3-5}\cline{7-10}
         \multicolumn{2}{c|}{}& Holder & Target & Exp & & UF\textsubscript{1} & LF\textsubscript{1} & NSF\textsubscript{1} & SF\textsubscript{1} \\ 
         \hline 
        \multicolumn{2}{|c|}{\cite{peng2021sparse}} & 65.8 & 71.0 & 76.7 & 59.6 & & 66.1 & - & 62.7 \\
        \multicolumn{2}{|c|}{\cite{samuel-etal-2022-direct}\textsuperscript{NC}} & 58.9 & 63.5 & 73.9 & - & - & - & 59.8 & 58.6 \\
        \multicolumn{2}{|c|}{\cite{samuel-etal-2022-direct}\textsuperscript{LE}} &57.6 & 64.9 & 72.5 & - & - & - & 60.0 & 58.8 \\
        \multicolumn{2}{|c|}{\cite{samuel-etal-2022-direct}\textsuperscript{OT}} &  64.2 & 67.4 & 73.2 & - & - & - & 62.5 & 61.3 \\
        \multicolumn{2}{|c|}{\cite{li2025extracting}} & 61.4 & - & 70.0 & 74.5 & - & - & 63.8 & 62.6 \\
        \hline
        \parbox[t]{2mm}{\multirow{13}{*}{\rotatebox[origin=c]{90}{\textit{h.first}}}} 
            & \textbf{A} & 70.3 & 76.3 & 80.2 & 57.3 & 57.9 & 56.5 & 46.1 & 45.0 \\
            & \textbf{R} & 69.8 & 75.1 & 78.3 & 63.0 & 62.3 & 60.1 & 51.9 & 50.6 \\ 
            & \textbf{B}\textsubscript{1} & 67.8 & 71.3 & 76.2 & 62.8 & 62.7 & 60.5 & 48.3 & 47.6 \\
            & \textbf{B}\textsubscript{2} & 72.1 & 71.8 & \textbf{80.9} & 64.5 & 66.0 & 64.0 & 50.4 & 49.6 \\
            & \textbf{B}\textsubscript{3} & 71.8 & 74.2 & \textbf{80.9} & 67.1 & 66.4 & 64.2 & 54.8 & 52.9 \\ 
            & \textbf{B6}\textsubscript{1} & 69.9 & 77.7 & 82.7 & \textbf{71.0} & \textbf{70.7} & \textbf{68.9} & 57.2 & 56.8\\
            & \textbf{B6}\textsubscript{2} & 69.6 & 76.2 & 80.2 & 70.7 & 69.6 & 67.5 & 61.9 & 60.6 \\
            & \textbf{B6}\textsubscript{3} & 72.2 & 75.8 & 80.7 & 69.2 & 70.1 & 68.5 & 62.1 & 61.3\\
            \cline{2-10}
            & \cite{barnes-etal-2021-structured} & 60.4 & 64.0 & 73.9 & 57.8 & 64.6 & 60.0 & 58.0 & 54.7 \\
            & \cite{fernandez2026structured} & 67.0 & 67.7 & 77.2 & 59.1 & 68.2 & 63.4 & 63.7 & 60.2 \\
            & Biaf & 66.7 & 70.6 & 77.4 & 50.0 & 48.5 & 46.8 & 40.8 & 39.7 \\ 
        \hline 
        \parbox[t]{2mm}{\multirow{13}{*}{\rotatebox[origin=c]{90}{\textit{h.final}}}}  
            & \textbf{A} &  66.1 & 71.3 & 77.9 & 50.2 & 47.8 & 46.2 & 37.9 & 37.1\\
            & \textbf{R} &  73.2 & 75.2 & 79.4 & 62.3 & 61.5 & 58.9 & 50.9 & 49.6  \\
            & \textbf{B}\textsubscript{1} & 65.5 & 72.8 & 74.7 & 56.6 & 64.4 & 62.3 & 50.7 & 49.7 \\ 
            & \textbf{B}\textsubscript{2} & \textbf{74.3} & 70.9 & 77.8 & 60.9 & 67.8 & 65.7 & 61.0 & 59.9 \\
            & \textbf{B}\textsubscript{3} & 70.1 & 69.1 & 77.8 & 61.0 & 66.9 & 64.4 & 60.9 & 59.4 \\ 
            & \textbf{B6}\textsubscript{1} & 69.8 & \textbf{77.8} & 82.5 & 64.5 & 67.2 & 64.5 & 57.8 & 56.5 \\
            & \textbf{B6}\textsubscript{2} & 71.8 & 74.4 & 79.8 & 66.0 & 68.0 & 65.6 & 63.7 & 62.6 \\
            & \textbf{B6}\textsubscript{3} & 71.8 & 76.1 & 78.9 & 66.8 & 66.5 & 64.3 & \textbf{65.1} & \textbf{63.8} \\
            \cline{2-10}
            & \cite{barnes-etal-2021-structured} & 60.5 & 64.0 & 72.1 & 56.9 & 60.8 & 56.0 & 58.0 & 54.7 \\
            & \cite{fernandez2026structured} & 66.6 & 67.9 & 76.6 & 60.5 & 66.9 & 61.4 & 63.8 & 60.4 \\
            & Biaf & 71.4 & 74.4 & 78.2 & 55.0 & 57.0 & 55.6 & 44.4 & 44.0\\ 
        \hline 
    \end{tabular}%
    }
\end{table}

\begin{table}[h]\centering\footnotesize
    \caption{Fine-grained results on the NoReC dataset.}\label{tab:norec-results}
    \resizebox{\columnwidth}{!}{%
    \setlength{\tabcolsep}{4pt}
    \renewcommand{\arraystretch}{1.1}
    \begin{tabular}{|l|c|ccc|c|cc|cc|}
        \cline{3-10} 
         \multicolumn{2}{c|}{}& \multicolumn{3}{c|}{\textbf{Spans (F\textsubscript{1})}} & \multirow{2}{*}{\textbf{T.SF\textsubscript{1}}} & \multicolumn{2}{c}{\textbf{Dep.}} & \multicolumn{2}{c|}{\textbf{Sent.}} \\ 
         \cline{3-5}\cline{7-10}
         \multicolumn{2}{c|}{}& Holder & Target & Exp & & UF\textsubscript{1} & LF\textsubscript{1} & NSF\textsubscript{1} & SF\textsubscript{1} \\ 
         \hline 
        \multicolumn{2}{|c|}{\cite{peng2021sparse}} & 63.6 & 55.3 & 56.1 &  31.5 & &  40.0 & - & 32.7 \\
        \multicolumn{2}{|c|}{\cite{samuel-etal-2022-direct}\textsuperscript{NC}} & 60.3 & 51.8 & 54.2 & - & - & - & 42.7 & 39.3 \\
        \multicolumn{2}{|c|}{\cite{samuel-etal-2022-direct}\textsuperscript{LE}} & 64.0 & 52.3 & 56.1 & - & - & - & 43.7 & 40.4 \\
        \multicolumn{2}{|c|}{\cite{samuel-etal-2022-direct}\textsuperscript{OT}} & 65.1 & 58.3 & 60.7 & - & - & - & \textbf{47.8} & 41.6 \\
        \multicolumn{2}{|c|}{\cite{li2025extracting}} & 62.5 & - & 61.4 & 61.3 & - & -  & 46.7 &  41.9 \\
        \hline
        \parbox[t]{2mm}{\multirow{13}{*}{\rotatebox[origin=c]{90}{\textit{h.first}}}} 
            & \textbf{A} &  57.4 & 60.6 & 62.3 & 35.4 & 29.6 & 25.9 & 27.1 & 25.1 \\ 
            & \textbf{R} & 67.2 & 62.7 & 63.1 & 42.6 & 42.1 & 37.6 & 29.6 & 27.5 \\
            & \textbf{B}\textsubscript{1} & 50.5 & 58.1 & 62.3 & 37.7 & 40.5 & 36.6 & 28.1 & 24.5 \\
            & \textbf{B}\textsubscript{2} & 50.9 & 59.3 & 64.3 & 44.9 & 40.8 & 36.4 & 32.4 & 29.3 \\ 
            & \textbf{B}\textsubscript{3} & 52.0 & 62.2 & 64.3 & 49.6 & 40.7 & 36.4 & 29.5 & 26.4 \\ 
            & \textbf{B6}\textsubscript{1} & 58.9 & 60.5 & 64.5 & 50.0 & 45.0 & 40.2 & 35.5 & 32.3\\
            & \textbf{B6}\textsubscript{2} & 56.9 & 62.8 & 54.2 & 28.4 & 51.9 & 50.1 & 28.8 & 23.9 \\
            & \textbf{B6}\textsubscript{3} & 57.8 & 62.5 & 63.9 & \textbf{52.7} & 44.9 & 40.2 & 40.6 & 37.6\\
            \cline{2-10}
            & \cite{barnes-etal-2021-structured} & 51.1 & 50.1 & 54.4 & 30.5 & 39.2 & 31.5 & 37.0 & 29.5  \\
            & \cite{fernandez2026structured} & 55.8 & 53.8 & 52.0 & 41.4 & 39.9 & 32.8 & 43.6 & 35.8 \\
            & Biaf &  56.0 & 60.4 & 59.1 & 26.0 & 27.1 & 24.9 & 23.4 & 21.5\\ 
        \hline 
        \parbox[t]{2mm}{\multirow{13}{*}{\rotatebox[origin=c]{90}{\textit{h.final}}}}  
            & \textbf{A} &  59.2 & 59.2 & 59.9 & 24.4 & 28.1 & 25.6 & 22.7 & 20.5  \\
            & \textbf{R} & 59.7 & 57.1 & 56.4 & 32.5 & 45.7 & 41.14 & 24.1 & 22.5 \\
            & \textbf{B}\textsubscript{1} & 55.2 & 57.9 & 58.8 & 32.4 & 51.7 & 45.8 & 30.9 & 27.9\\
            & \textbf{B}\textsubscript{2} &  54.5 & 59.4 & 61.4 & 38.9 & 53.3 & 48.0 & 41.3 & 37.3  \\
            & \textbf{B}\textsubscript{3} & 55.6 & 59.3 & 62.8 & 39.4 & 53.8 & 47.7 & 42.0 & 37.9 \\ 
            & \textbf{B6}\textsubscript{1} & \textbf{67.7} & \textbf{63.7} & 63.9 & 47.3 & 54.9 & 49.5 & 40.4 & 37.0\\
            & \textbf{B6}\textsubscript{2} & 58.4 & 63.4 & 64.7 & 47.4 & \textbf{56.6} & \textbf{51.4} & 46.8 & 42.5\\
            & \textbf{B6}\textsubscript{3} & 62.8 & 62.0 & \textbf{66.1} & 47.5 & 55.9 & 50.5 & 47.4 & \textbf{43.8}\\
            \cline{2-10}
            & \cite{barnes-etal-2021-structured} & 60.4 & 54.8 & 55.5 & 31.9 & 48.0 & 37.7 & 39.2 & 31.2 \\
            & \cite{fernandez2026structured} & 57.5 & 54.3 & 51.5 & 36.1 & 45.0 & 36.5 & 44.2 & 35.8 \\
            & Biaf & 64.2 & 61.8 & 61.3 & 33.8 & 30.3 & 27.3 & 28.2 & 26.6 \\ 
        \hline 
    \end{tabular}%
    }
\end{table}

\begin{table}[h]\centering\footnotesize
    \caption{Fine-grained results on the Darmstadt Universities dataset. Same notation as in Table \ref{tab:mb-ca-results}.}\label{tab:ds-results}
    \resizebox{\columnwidth}{!}{%
    \setlength{\tabcolsep}{4pt}
    \renewcommand{\arraystretch}{1.1}
    \begin{tabular}{|l|c|ccc|c|cc|cc|}
        \cline{3-10} 
         \multicolumn{2}{c|}{}& \multicolumn{3}{c|}{\textbf{Spans (F\textsubscript{1})}} & \multirow{2}{*}{\textbf{T.F\textsubscript{1}}} & \multicolumn{2}{c|}{\textbf{Dep.}} & \multicolumn{2}{c|}{\textbf{Sent.}} \\ 
         \cline{3-5}\cline{7-10}
         \multicolumn{2}{c|}{}& Holder & Target & Exp. & & UF\textsubscript{1} & LF\textsubscript{1} & NSF\textsubscript{1} & SF\textsubscript{1} \\ 
         \hline 
        \multicolumn{2}{|c|}{\cite{peng2021sparse}} & 50.0 & 44.8 & 43.7 & 30.7 & - & 34.9 & - & 27.4 \\
        \multicolumn{2}{|c|}{\cite{samuel-etal-2022-direct}\textsuperscript{NC}} &  31.4 & 35.0 & 35.1 & - & - & - & 24.8 & 22.9 \\
        \multicolumn{2}{|c|}{\cite{samuel-etal-2022-direct}\textsuperscript{LE}} &32.5 & 38.0 & 36.2 & - & - & - & 28.8 & 27.3  \\
        \multicolumn{2}{|c|}{\cite{samuel-etal-2022-direct}\textsuperscript{OT}} &  42.2 & 40.6 & 39.3 & - & - & - & 33.2 & 31.2\\
        \multicolumn{2}{|c|}{\cite{li2025extracting}} & 47.0 & - & 44.7 & 45.0 & - & - &  37.8 & 34.1 \\
        \hline
        \parbox[t]{2mm}{\multirow{13}{*}{\rotatebox[origin=c]{90}{\textit{h.first}}}} 
            & \textbf{A} & 52.9 & 45.9 & 42.3 & 24.6 & 23.3 & 21.1 & 16.7 & 15.3 \\
            & \textbf{R} & 52.6 & 46.2 & 44.3 & 25.8 & 24.0 & 21.3 & 17.3 & 16.2 \\
            & \textbf{B}\textsubscript{1} & 37.5 & 43.8 & 50.8 & 36.7 & 38.2 & 36.4 & 29.0 & 27.8 \\ 
            & \textbf{B}\textsubscript{2} & 47.2 & 45.0 & 48.4 & 31.5 & 37.8 & 35.3 & 27.6 & 23.8 \\ 
            & \textbf{B}\textsubscript{3} & 47.6 & 45.0 & 48.4 & 31.5 & 37.8 & 35.3 & 27.6 & 23.8 \\
            & \textbf{B6}\textsubscript{1} & 47.1 & 49.0 & 50.0 & 40.0 & 42.1 & 40.1 & 37.3 & \textbf{34.8} \\
            & \textbf{B6}\textsubscript{2} & 37.5 & 47.7 & 51.3 & \textbf{41.1} & 42.6 & 40.4 & 36.5 & 34.2 \\
            & \textbf{B6}\textsubscript{3} & 35.3 & 51.4 & \textbf{53.7} & 39.6 & 45.3 & 42.6 & \textbf{39.2} & \textbf{34.8} \\
            \cline{2-10}
            & \cite{barnes-etal-2021-structured} & 28.0 & 39.9 & 40.3 & 26.7 & 35.3 & 31.4 & 31.0 & 25.0  \\
            & \cite{fernandez2026structured} & 40.4 & 44.3 & 44.3 & 32.5 & 37.3 & 32.5 & 35.9 & 29.9 \\
            & Biaf  & 40.0 & 33.9 & 45.6 & 16.6 & 26.6 & 24.9 & 8.3 & 8.3\\ 
        \hline 
        \parbox[t]{2mm}{\multirow{13}{*}{\rotatebox[origin=c]{90}{\textit{h.final}}}}  
            & \textbf{A} & 47.4 & 43.9 & 40.7 & 25.6 & 25.2 & 23.6 & 11.4 & 10.2 \\
            & \textbf{R} & 47.1 & 43.7 & 40.4 & 25.2 & 25.0 & 23.5 & 11.2 & 10.0 \\
            & \textbf{B}\textsubscript{1} & 47.1 & 46.6 & 45.4 & 32.3 & 39.9 & 36.3 & 33.2 & 27.0 \\ 
            & \textbf{B}\textsubscript{2} &  \textbf{63.2} & 44.2 & 45.2 & 34.9 & 39.5 & 38.2 & 31.8 & 30.3 \\ 
            & \textbf{B}\textsubscript{3} & 44.4 & 48.9 & 48.5 & 37.3 & 43.0 & 39.4 & 35.8 & 33.0 \\
            & \textbf{B6}\textsubscript{1} & 55.6 & 46.7 & 48.4 & 35.6 & 43.4 & 39.9 & 37.4 & 33.1 \\
            & \textbf{B6}\textsubscript{2} & 33.3 & \textbf{51.8} & 48.8 & 37.2 & 40.5 & 38.7 & 35.2 & 32.8 \\
            & \textbf{B6}\textsubscript{3} & 28.6 & 45.6 & 53.0 & 34.3 & \textbf{45.8} & \textbf{42.7} & 36.1 & 32.0\\
            \cline{2-10}
            & \cite{barnes-etal-2021-structured} &  37.4 & 42.1 & 45.5 & 29.6 & 38.1 & 33.9 & 34.3 & 26.5\\
            & \cite{fernandez2026structured} & 43.2 & 44.4 & 43.6 & 28.7 & 36.3 & 30.3 & 32.7 & 25.6 \\
            & Biaf & 37.5 & 35.3 & 43.8 & 18.0 & 23.4 & 21.7 & 9.7 & 9.7 \\ 
        \hline 
    \end{tabular}%
    }
\end{table}

\begin{table}[h]\centering\footnotesize
    \caption{Fine-grained results on the MPQA dataset.}\label{tab:mpqa-results}
    \resizebox{\columnwidth}{!}{%
    \setlength{\tabcolsep}{4pt}
    \renewcommand{\arraystretch}{1.1}
    \begin{tabular}{|l|c|ccc|c|cc|cc|}
        \cline{3-10} 
         \multicolumn{2}{c|}{}& \multicolumn{3}{c|}{\textbf{Spans (F\textsubscript{1})}} & \multirow{2}{*}{\textbf{T.SF\textsubscript{1}}} & \multicolumn{2}{c}{\textbf{Dep.}} & \multicolumn{2}{c|}{\textbf{Sent.}} \\ 
         \cline{3-5}\cline{7-10}
         \multicolumn{2}{c|}{}& Holder & Target & Exp & & UF\textsubscript{1} & LF\textsubscript{1} & NSF\textsubscript{1} & SF\textsubscript{1} \\ 
         \hline 
        \multicolumn{2}{|c|}{\cite{peng2021sparse}} & 47.9 & 50.7 & 47.8 & 33.7 & & 37.6 & - & 20.4  \\
        \multicolumn{2}{|c|}{\cite{samuel-etal-2022-direct}\textsuperscript{NC}} & 58.4 & 60.3 & 55.8 & - & - & - & 38.7 & 28.3  \\
        \multicolumn{2}{|c|}{\cite{samuel-etal-2022-direct}\textsuperscript{LE}} &  53.6 & 53.4 & 53.4 & - & - & - & 33.8 & 27.0 \\
        \multicolumn{2}{|c|}{\cite{samuel-etal-2022-direct}\textsuperscript{OT}} &  55.7 & 64.0 & 53.5 & - & - & - & \textbf{45.1} & \textbf{34.1}  \\
        \multicolumn{2}{|c|}{\cite{li2025extracting}} & 46.5 & - & 51.8 & 47.6 & - & - & 29.3 & 24.1 \\
        \hline
        \parbox[t]{2mm}{\multirow{13}{*}{\rotatebox[origin=c]{90}{\textit{h.first}}}} 
            & \textbf{A} & 54.2 & 59.0 & 53.3 & 32.2 & 23.1 & 21.6 & 11.8 & 9.5 \\
            & \textbf{R} & 55.5 & 64.2 & 45.7 & 35.9 & 37.9 & 36.8 & 17.9 & 15.7\\
            & \textbf{B}\textsubscript{1} & 55.0 & 61.9 & 46.1 & 41.4 & 40.6 & 38.9 & 10.7 &  9.3 \\
            & \textbf{B}\textsubscript{2} & 57.8 & 58.1 & 38.8 & 41.4 & 39.1 & 37.5 & 11.6 & 10.1 \\
            & \textbf{B}\textsubscript{3} & 56.9 & 57.3 & 49.6 & 42.6 & 42.0 & 40.3 & 17.9 & 15.4 \\ 
            & \textbf{B6}\textsubscript{1} & 57.8 & 62.9 & 51.9 & 44.2 & 49.3 & 47.3 & 30.8 & 26.5 \\
            & \textbf{B6}\textsubscript{2} & 53.7 & \textbf{64.4} & 50.3 & 49.1 & 47.1 & 45.5 & 24.9 & 20.9\\
            & \textbf{B6}\textsubscript{3} & 57.0 & 64.2 & 56.3 & 50.2 & 49.1 & 47.3 & 23.2 & 19.3\\
            \cline{2-10}
            & \cite{barnes-etal-2021-structured} & 43.8 & 51.0 & 48.1 & 33.5 & 40.0 & 36.9 & 24.5 & 17.4 \\
            & \cite{fernandez2026structured} & 46.2 & 49.7 & 44.7 & 33.9 & 40.5 & 37.6 & 28.7 & 22.2 \\
            & Biaf & 52.2 & 61.2 & 52.6 & 8.2 & 18.7 & 17.2 & 9.1 & 7.8\\ 
        \hline 
        \parbox[t]{2mm}{\multirow{13}{*}{\rotatebox[origin=c]{90}{\textit{h.final}}}}  
            & \textbf{A} & 56.2 & 53.6 & 52.1 & 9.5  & 18.0 & 16.8 & 11.1 & 9.8  \\
            & \textbf{R} & 56.4 & 61.4 & 49.4 & 17.0 & 38.3 & 37.0 & 8.5  & 8.2\\
            & \textbf{B}\textsubscript{1} & 54.3 & 53.0 & 50.7 & 20.6 & 44.4 & 42.5 & 22.1 & 18.6 \\  
            & \textbf{B}\textsubscript{2} &  39.4 & 45.0 & 37.7 & 16.7 & 34.7 & 33.8 & 9.9 & 8.2 \\ 
            & \textbf{B}\textsubscript{3} & 55.7 & 56.0 & \textbf{56.2} & 20.5 & 47.9 & 45.9 & 23.4 & 19.8 \\ 
            & \textbf{B6}\textsubscript{1} & \textbf{60.0} & 62.5 & 54.1 & 28.0 & 50.8 & 49.0 & 32.1 & 26.8 \\
            & \textbf{B6}\textsubscript{2} & 56.9 & 62.8 & 54.2 & 28.4 & \textbf{51.9} & \textbf{50.1} & 28.8 & 23.9 \\
            & \textbf{B6}\textsubscript{3} & 55.6 & 61.0 & 54.5 & 26.0 & 50.2 & 47.8 & 29.9 & 23.3 \\
            \cline{2-10}
            & \cite{barnes-etal-2021-structured} & 46.3 &   49.5 &  46.0  & 18.6 &  41.4 &  38.0  & 26.1 &  18.8 \\
            & \cite{fernandez2026structured} & 43.8 & 46.4 & 45.2 & 18.1 & 38.0 & 35.4 & 25.9 & 19.7 \\
            & Biaf & 52.4 & 55.5 & 53.0 & 29.0 & 25.2 & 24.0 & 14.0 & 12.0\\ 
        \hline 
    \end{tabular}%
    }
\end{table}

\begin{table}[h]\centering\footnotesize
    \caption{Fine-grained results on the OpeNER (English) dataset.}\label{tab:opener-en-results}
    \resizebox{\columnwidth}{!}{%
    \setlength{\tabcolsep}{4pt}
    \renewcommand{\arraystretch}{1.1}
    \begin{tabular}{|l|c|ccc|c|cc|cc|}
        \cline{3-10} 
         \multicolumn{2}{c|}{}& \multicolumn{3}{c|}{\textbf{Spans (F\textsubscript{1})}} & \multirow{2}{*}{\textbf{T.SF\textsubscript{1}}} & \multicolumn{2}{c}{\textbf{Dep.}} & \multicolumn{2}{c|}{\textbf{Sent.}} \\ 
         \cline{3-5}\cline{7-10}
         \multicolumn{2}{c|}{}& Holder & Target & Exp & & UF\textsubscript{1} & LF\textsubscript{1} & NSF\textsubscript{1} & SF\textsubscript{1} \\ 
         \hline 
        \parbox[t]{2mm}{\multirow{13}{*}{\rotatebox[origin=c]{90}{\textit{h.first}}}} 
            & \textbf{A} &  66.2 & 77.7 & 76.2 & 53.0 & 50.0 & 49.0 & 41.8 & 41.2 \\ 
            & \textbf{R} & 70.4 & 81.9 & 74.3 & 69.4 & 63.4 & 61.4 & 53.8 & 52.5 \\
            & \textbf{B}\textsubscript{1} & \textbf{73.1} & 80.9 & 75.3 & 73.8 & 66.8 & 65.2 & 51.7 & 51.3 \\
            & \textbf{B}\textsubscript{2} & 68.8 & \textbf{82.3} & 76.9 & 75.6 & 68.6 & 66.7 & 56.1 & 55.0  \\
            & \textbf{B}\textsubscript{3} & 65.7 & 80.8 & 75.6 & 75.8 & 67.4 & 65.6 & 54.0 & 53.0\\
            & \textbf{B6}\textsubscript{1} & 72.2 & 81.9 & \textbf{78.4} & 74.7 & 68.9 & 67.2 & 58.7 & 57.9\\
            & \textbf{B6}\textsubscript{2} & \textbf{73.1} & 81.6 & 78.2 & \textbf{76.2} & \textbf{71.4} & \textbf{69.6} & 66.0 & 65.1 \\
            & \textbf{B6}\textsubscript{3} & 69.7 & 81.2 & 77.3 & 74.6 & 69.8 & 68.1 & 65.2 & 64.0\\
            \cline{2-10}
            & Biaf & 71.6 & 76.8 & 76.2 & 48.8 & 46.0 & 44.9 & 36.9 & 35.9 \\ 
        \hline 
        \parbox[t]{2mm}{\multirow{13}{*}{\rotatebox[origin=c]{90}{\textit{h.final}}}}  
            & \textbf{A} & 65.0 & 77.0 & 74.8 & 48.3 & 47.9 & 46.8 & 37.6 & 36.6 \\ 
            & \textbf{R} & 70.1 & 80.7 & 73.8 & 65.1 & 61.5 & 59.4 & 48.1 & 47.1 \\ 
            & \textbf{B}\textsubscript{1} & 56.1 & 79.2 & 73.4 & 70.0 & 68.8 & 67.0 & 58.8 & 58.0\\
            & \textbf{B}\textsubscript{2} & 63.5 & 79.9 & 77.2 & 71.4 & 69.7 & 67.9 & 66.3 & 64.6 \\
            & \textbf{B}\textsubscript{3} & 58.3 & 78.5 & 76.3 & 69.0 & 69.2 & 66.8 & 64.7 & 63.0 \\
            & \textbf{B6}\textsubscript{1} & 71.1 & 81.4 & 78.3 & 72.6 & 69.8 & 67.9 & 60.8 & 59.6 \\
            & \textbf{B6}\textsubscript{2} & 61.6 & 82.0 & 77.7 & 74.0 & 70.0 & 68.2 & 67.9 & 66.6 \\
            & \textbf{B6}\textsubscript{3} & 70.0 & 80.9 & 77.9 & 73.0 & 70.6 & 68.5 & \textbf{70.8} & \textbf{69.8} \\
            \cline{2-10}
            & Biaf & 65.7 & 72.1 & 73.0 & 52.5 & 50.0 & 48.6 & 41.9 & 41.4 \\ 
        \hline 
    \end{tabular}%
    }
\end{table}

\begin{table}[h]\centering\footnotesize
    \caption{Fine-grained results on the OpeNER (Spanish) dataset.}\label{tab:opener-es-results}
    \resizebox{\columnwidth}{!}{%
    \setlength{\tabcolsep}{4pt}
    \renewcommand{\arraystretch}{1.1}
    \begin{tabular}{|l|c|ccc|c|cc|cc|}
        \cline{3-10} 
         \multicolumn{2}{c|}{}& \multicolumn{3}{c|}{\textbf{Spans (F\textsubscript{1})}} & \multirow{2}{*}{\textbf{T.SF\textsubscript{1}}} & \multicolumn{2}{c}{\textbf{Dep.}} & \multicolumn{2}{c|}{\textbf{Sent.}} \\ 
         \cline{3-5}\cline{7-10}
         \multicolumn{2}{c|}{}& Holder & Target & Exp & & UF\textsubscript{1} & LF\textsubscript{1} & NSF\textsubscript{1} & SF\textsubscript{1} \\ 
         \hline 
        \parbox[t]{2mm}{\multirow{13}{*}{\rotatebox[origin=c]{90}{\textit{h.first}}}} 
            & \textbf{A} &  68.4 & 74.7 & 75.2 & 48.9 & 45.8 & 44.5 & 39.0 & 38.5 \\
            & \textbf{R} & 75.9 & 79.7 & 71.3 & 63.7 & 58.5 & 56.4 & 47.0 & 45.6 \\
            & \textbf{B}\textsubscript{1} & 67.5 & 78.1 & 73.1 & 68.6 & 61.8 & 60.4 & 49.0 & 47.9 \\
            & \textbf{B}\textsubscript{2} & 70.6 & 79.7 & 77.0 & 71.5 & 64.8 & 62.9 & 57.7 & 56.5 \\
            & \textbf{B}\textsubscript{3}  & 69.9 & 79.1 & 75.2 & 70.3 & 64.5 & 62.9 & 56.4 & 55.4\\ 
            & \textbf{B6}\textsubscript{1} & 72.9 & 79.9 & 76.7 & 72.0 & 65.6 & 63.3 & 56.7 & 55.3 \\
            & \textbf{B6}\textsubscript{2} & 70.5 & 81.1 & \textbf{78.1} & \textbf{73.8} & 67.1 & 65.3 & 64.1 & 62.6 \\
            & \textbf{B6}\textsubscript{3} & 70.3 & 80.4 & 77.8 & 72.9 & 66.0 & 63.8 & 64.0 & 62.1\\
            \cline{2-10}
            & Biaf & 61.11 & 73.00 & 74.78 & 43.8 & 43.8 & 42.9 & 34.1 & 33.9 \\ 
        \hline 
        \parbox[t]{2mm}{\multirow{13}{*}{\rotatebox[origin=c]{90}{\textit{h.final}}}}  
            & \textbf{A} & 69.0 & 76.4 & 74.2 & 44.9 & 44.6 & 43.3 & 36.3 & 35.9\\ 
            & \textbf{R} & 80.5 & 79.1 & 70.9 & 60.1 & 61.6 & 58.7 & 44.1 & 43.3 \\
            & \textbf{B}\textsubscript{1} & 78.7 & 78.8 & 73.0 & 65.7 & 69.1 & 66.5 & 61.8 & 60.6 \\
            & \textbf{B}\textsubscript{2} & 75.9 & 79.9 & 76.5 & 67.9 & 71.0 & \textbf{68.8} & 66.8 & 65.4 \\ 
            & \textbf{B}\textsubscript{3} & 74.4 & 77.9 & 76.4 & 65.2 & 70.6 & 67.8 & 63.7 & 62.1 \\ 
            & \textbf{B6}\textsubscript{1} & 73.7 & 81.0 & 77.3 & 69.3 & 70.3 & 67.1 & 62.0 & 60.4\\
            & \textbf{B6}\textsubscript{2} & \textbf{79.1} & \textbf{81.8} & 77.6 & 70.0 & \textbf{71.3} & 68.7 & 67.6 & 66.0\\
            & \textbf{B6}\textsubscript{3} &  74.4 & 78.8 & 76.5 & 68.7 & 69.2 & 66.8 & \textbf{68.7} & \textbf{67.6} \\
            \cline{2-10}
            & Biaf & 56.3 & 72.4 & 72.5 & 48.3 & 44.4 & 43.1 & 39.6 & 39.0 \\ 
        \hline 
    \end{tabular}%
    }
\end{table}

\begin{table}[ht!]\centering\footnotesize
\caption{Model efficiency in terms of training and inference throughput (tokens/s, sentences/s)}
\label{tab:model-efficiency-analysis}
\renewcommand{\arraystretch}{1.2}
\setlength{\tabcolsep}{5pt}
\begin{tabular}{|l|cc|cc|}
\hline
\multirow{2}{*}{\textbf{Encoding}} & \multicolumn{2}{c|}{\textbf{Training}} & \multicolumn{2}{c|}{\textbf{Inference}} \\
\cline{2-5}
& \textbf{Token/s} & \textbf{Sentences/s} & \textbf{Token/s} & \textbf{Sentences/s} \\
\hline
\textbf{A} & 3792.25 & 224.78 & 4454.28 & 268.80 \\
\textbf{R} & 3818.01 & 225.91 & 4491.96 & 270.72 \\
\textbf{B}\textsubscript{1} & 3831.82 & 226.45 & 4409.86 & 266.07 \\
\textbf{B}\textsubscript{2} & 3796.78 & 224.57 & 4502.59 & 271.32 \\
\textbf{B}\textsubscript{3} & 3828.62 & 226.31 & 4486.74 & 270.80 \\
\textbf{B6}\textsubscript{1} & 3818.69 & 225.50 & 4461.88 & 269.35 \\
\textbf{B6}\textsubscript{2} & 3756.26 & 222.04 & 4367.45 & 263.62 \\
\textbf{B6}\textsubscript{3} & 3784.85 & 223.77 & 4378.02 & 263.75 \\
\hline
Biaff & 1892.87 & 111.46 & 2195.78 & 122.59 \\
\hline
\end{tabular}
\end{table}

\clearpage 

\end{document}